\documentclass[sigconf]{acmart}
\AtBeginDocument{%
  }

\setcopyright{acmlicensed}
\copyrightyear{2018}
\acmYear{2018}
\acmDOI{XXXXXXX.XXXXXXX}
\acmConference[Conference acronym 'XX]{Make sure to enter the correct
  conference title from your rights confirmation email}{June 03--05,
  2018}{Woodstock, NY}
\acmISBN{978-1-4503-XXXX-X/2018/06}

\usepackage{multirow,soul}
\usepackage[para]{footmisc}

\author{Sarmistha Das}
\affiliation{%
  \institution{Indian Institute of Technology Patna}
  \city{Patna}
  \country{India}}
  
\email{sarmistha1515@gmail.com}

\author{R E Zera Marveen Lyngkhoi}
\affiliation{%
  \institution{Indian Institute of Technology Patna}
  \city{Patna}
  \country{India}}
\email{zera\_2311ai06@iitp.ac.in}

\author{Kirtan Jain}
\affiliation{%
  \institution{Indian Institute of Technology Patna}
  \city{Patna}
  \country{India}}
\email{kirtanjain0504@gmail.com}

\author{Vinayak Goyal}
\affiliation{%
  \institution{Indian Institute of Technology Patna}
  \city{Patna}
  \country{India}}
\email{vinayakgoyal2410@gmail.com}

\author{Sriparna Saha}
\affiliation{%
  \institution{Indian Institute of Technology Patna}
  \city{Patna}
  \country{India}}
\email{sriparna@iitp.ac.in}

\author{Manish Gupta}
\affiliation{%
  \institution{Microsoft, India}
  \city{Bengaluru}
  \country{India}}
\email{gmanish@microsoft.com}

\renewcommand{\shortauthors}{Sarmistha Das et al.}
\copyrightyear{2025}
\acmYear{2025}
\setcopyright{cc}
\setcctype{by-nc-nd}

\acmConference[CIKM '25]{Proceedings of the 34th ACM International
Conference on Information and Knowledge Management}{November 10--14,
2025}{Seoul, Republic of Korea}
\acmBooktitle{Proceedings of the 34th ACM International Conference on
Information and Knowledge Management (CIKM '25), November 10--14, 2025,
Seoul, Republic of Korea}\acmDOI{10.1145/3746252.3761576}
\acmISBN{979-8-4007-2040-6/2025/11}
\begin{document}

\title{When Words Can't Capture It All: Towards Video-Based User Complaint Text Generation with Multimodal Video Complaint Dataset}
\renewcommand{\shorttitle}{Video-Based User Complaint Text Generation}
\begin{abstract}
While there exists a lot of work on explainable complaint mining, articulating user concerns through text or video remains a significant challenge, often leaving issues unresolved. Users frequently struggle to express their complaints clearly in text but can easily upload videos depicting product defects (e.g., vague text such as \textit{`worst product'} paired with a 5-second video depicting a broken headphone with the right earcup). This paper formulates a new task in the field of complaint mining to aid the common users' need to write an expressive complaint, which is Complaint Description from Videos (\textsc{CoD-V}) (e.g., to help the above user articulate her complaint about the defective right earcup). To this end, we introduce \textsc{ComVID}, a video complaint dataset containing 1,175 complaint videos and the corresponding descriptions, also annotated with the emotional state of the complainer. Additionally, we present a new complaint retention (CR) evaluation metric that discriminates the proposed (\textsc{CoD-V}) task against standard video summary generation and description tasks. To strengthen this initiative, we introduce a multimodal Retrieval-Augmented Generation (RAG) embedded VideoLLaMA2-7b model, designed to generate complaints while accounting for the user's emotional state. We conduct a comprehensive evaluation of several Video Language Models on several tasks (pre-trained and fine-tuned versions) with a range of established evaluation metrics, including METEOR, perplexity, and the Coleman-Liau readability score, among others. Our study lays the foundation for a new research direction to provide a platform for users to express complaints through video. Dataset and resources are available at: \url{https://github.com/sarmistha-D/CoD-V}.

\end{abstract}

\ccsdesc[500]{Computing methodologies~Natural language generation}
\ccsdesc[300]{Computing methodologies~Video summarization}
\ccsdesc[500]{Computing methodologies~Computer vision tasks}
\ccsdesc[300]{Computing methodologies~Neural networks}

\keywords{Video analysis, complaint generation, video to text, \textsc{ComVID}}


\maketitle

\section{Introduction}
In recent years, e-commerce giants have made significant strides in expanding their online marketing efforts, particularly targeting rural or underserved areas, e.g., Amazon, Jio Mart, and Flipkart in India\footnote{\href{https://timesofindia.indiatimes.com/business/india-business/retail-giants-eye-rural-india-small-towns-to-push-growth/articleshow/78713752.cms}{https://timesofindia/retail-giants-eye-rural-india-small-towns- to-push-growth}}, Jumia in Nigeria\footnote{\href{https://www.jumia.com.ng/}{jumia.com.ng}}, Mercado Libre in Latin America\footnote{\href{https://mercadolibre.com/}{mercadolibre.com}}, and Daraz in South Asia\footnote{\href{https://daraz.com/}{daraz.com}}. Comprehending customer mindset is essential for engineering advanced interactive complaint analysis systems that enable significant engagement, especially for individuals who find it difficult to express their grievances in written form.

\begin{figure}[!t]    
\centering
    \includegraphics[width=\columnwidth]{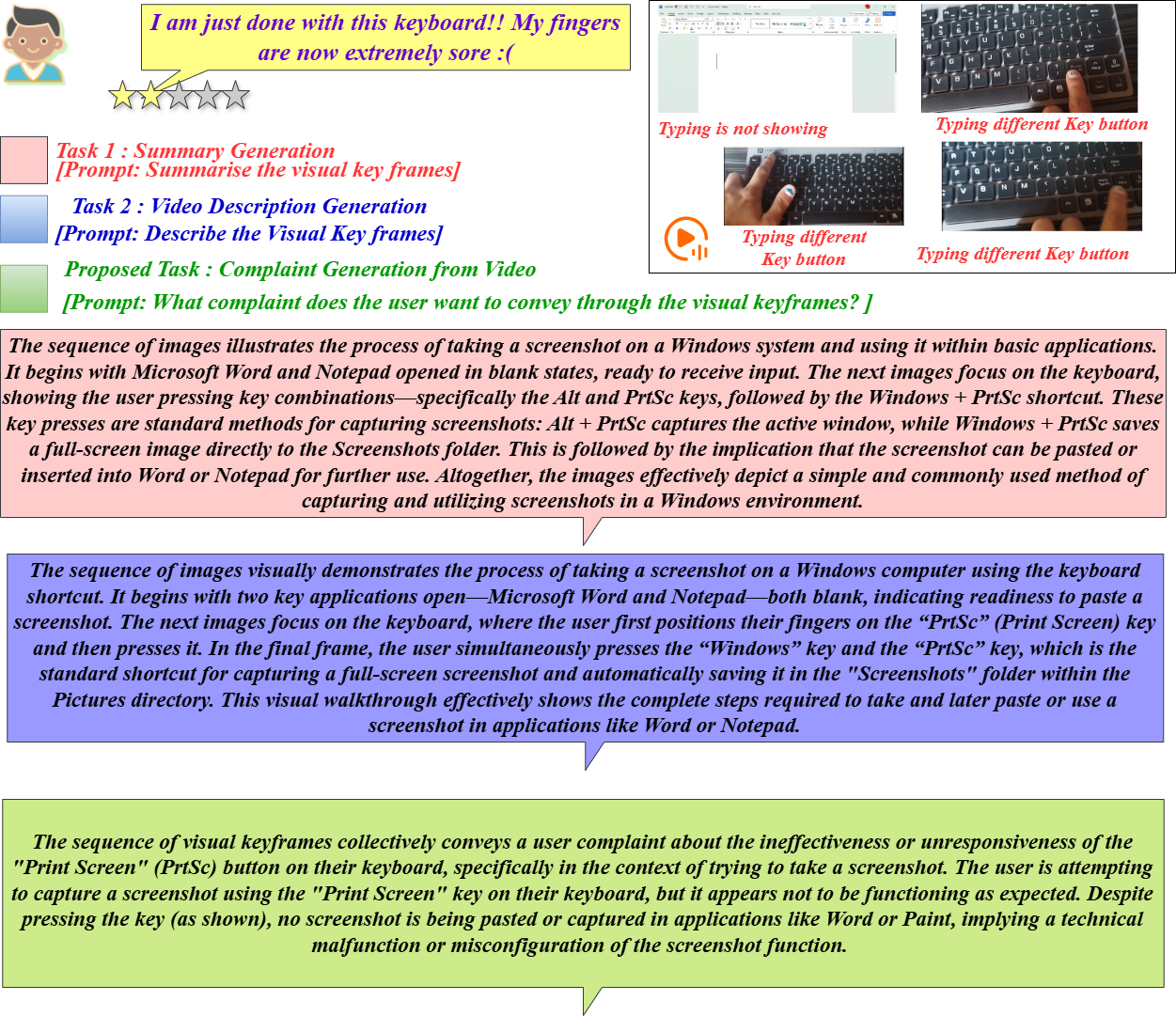}
    \caption{Comparative Analysis of Proposed Complaint Generation Task, Traditional Summary Generation, and Video Description Tasks on User-Uploaded Video Reviews}
    \label{intro}
\end{figure}

Effectively expressing complaints is crucial for building an inclusive marketplace, especially for individuals with expressive language disorders or limited education—such as farmers, transport laborers, and retail workers—who are increasingly shopping online. For organizations, customer reviews and ratings are vital for analyzing user sentiment toward products \cite{alqahtani2021product,wassan2021amazon}. However, users with limited literacy often struggle to convey nuanced issues, resulting in sparse or ambiguous complaints. To address this, prior work explainable complaint-mining \cite{das2024negative} used transformer-based models on Twitter data, combining sentence- and word-level encoders with attention-guided explanations to classify complaints and infer root causes. Building on this, a later study \cite{das2025deciphering} introduced a multimodal vision–language transformer with image-segment encoding to jointly process video features and transcripts, enabling precise mapping of nonverbal cues and spoken narratives to complaint causes.


Although many users can upload videos illustrating these issues, the impact is often diluted by unclear communication. For example, one user posted a video review showing broken headphones (see Fig. \ref{intro}) alongside a brief text review stating, \textit{``I am just done with this keyboard!!My fingers are now extremely sore''} expecting an \textit{exchange or repair}. Customer service teams struggle with high complaint volumes, while traditional video summarization models produce generic descriptions that miss the core issue, such as \textit{``the person is trying to take a screenshot''}. Fine-grained analysis tools also fail to detect implicit intent, delaying resolution.

\noindent\textbf{\textit{Motivation}}: Previous studies have focused on fine-grained multimodal complaint mining, examining specific aspects \cite{singh2023knowing,jain2023abcord}, but often fall short in capturing the full intent behind user complaints, such as specific issues like \textit{non-functioning keyboards} or \textit{poor product quality} highlighted through video content. This limitation poses significant challenges, especially for individuals with dysgraphia, who face cognitive barriers in articulating their thoughts in written form, or those with constrained linguistic repertoires, thereby hindering their ability to effectively communicate their concerns or make informed purchasing decisions. Addressing the pressing need for an integrative framework that links visual media with corresponding textual descriptions, we introduce a pioneering complaint generation task, \textsc{CoD-V}, in the domain of complaint mining, complemented by a Complaint Nature Retention (CR) metric to substantiate our approach. This task seeks to generate detailed complaint narratives from user-uploaded videos, enhancing customers’ ability to convey their concerns and enabling a more nuanced understanding of their discomforting experiences.\\
\textbf{\textit{Research Objectives:}} 1) Our primary objective is to establish a structured platform that enables inarticulate or busy users to articulate their grievances through video content. We also aim to empower inarticulate consumers and enhance their engagement with e-commerce platforms. 2) We intend to investigate whether the generated textual descriptions from complaint videos effectively capture the factual elements and the emotional state accurately reflecting the user's true feelings and actual expectations. 3) Furthermore, we aim to elucidate how the proposed video‐based complaint generation task fundamentally differs from conventional video summarization and description tasks.
\\
\textbf{\textit{Contributions:}} We summarize the key contributions as follows: 1) We introduce a novel task, Complaint Description from Videos (\textsc{CoD-V}), along with the Complaint Retention (CR) metric, designed to generate and evaluate comprehensive complaint narratives directly from video content. This approach addresses a critical gap in existing methods for managing consumer grievances. 2) In addition, we present the unique \textsc{ComVID} dataset, which serves as an invaluable resource for researchers and practitioners to advance studies in video-based complaint analysis and generation. 3) Furthermore, we propose a fine-tuned Retrieval Augmented Generation (RAG)-embedded VideoLLaMA2-7b model and conduct extensive experiments on various video-language models. These experiments explore the impact of incorporating the complainer's emotional state, using several quantitative metrics for evaluation.
\section{Related Work}
The expansion of e-commerce to less literate demographics necessitates robust complaint-handling frameworks \cite{huang2024research,helmy2024decision,akin2024enhancing}. Recent advancements leverage transformers-based architectures for complaint severity classification \cite{jin2021modeling,10379488,singh2022sentiment}, while multimodal approaches incorporate vision-language integration for nuanced emotion and sentiment analysis \cite{poria2018meld,saha2021towards,singh2023knowing}. CMA-CLIP \cite{liu2021cma} employs a multimodal classifier for predefined attribute prediction but struggles with generalization to unseen data. Generative approaches~\cite{roy2021attribute,das2023let} reformulate classification as sequence generation task. 
\begin{table}[t]
 \caption{Comparison of complaint datasets and their associated labels.}\label{tab:resource}
\centering
\tabcolsep2pt
\scriptsize
\begin{tabular}{l|cccc|c}
\hline
{Datasets}&\multicolumn{4}{c|}{Labels}& Modality \\ \cline{2-6} 
& \multicolumn{1}{c|}{Complaint}&\multicolumn{1}{c|}{Emotion}&\multicolumn{1}{c|}{Description}&\multicolumn{1}{c|}{Count}&Text/Image/Video \\ \hline
Complaints \cite{DBLP:conf/acl/Preotiuc-Pietro19a}&\multicolumn{1}{c|}{$\checkmark$}&\multicolumn{1}{c|}{$\times$}&\multicolumn{1}{c|}{$\times$}&3499&Text \\ \hline
Complaint\_ESS \cite{singh2022adversarial}&\multicolumn{1}{c|}{$\checkmark$}&\multicolumn{1}{c|}{$\checkmark$}&\multicolumn{1}{c|}{$\times$}&1971&Text \\ \hline
CESAMARD \cite{singh2023knowing}&\multicolumn{1}{c|}{$\checkmark$}&\multicolumn{1}{c|}{$\checkmark$}&\multicolumn{1}{c|}{$\times$}&3962&Text + Image \\ \hline
X-FinCORP \cite{das2023let}&\multicolumn{1}{c|}{$\checkmark$}&\multicolumn{1}{c|}{$\checkmark$}&\multicolumn{1}{c|}{$\times$}&6282&Text \\ \hline
VCD \cite{devanathan2024seeing}&\multicolumn{1}{c|}{$\checkmark$}&\multicolumn{1}{c|}{$\times$}&\multicolumn{1}{c|}{$\times$}&450&Video \\ \hline
\textit{\textsc{ComVID} (Ours)}&\multicolumn{1}{c|}{$\checkmark$}&\multicolumn{1}{c|}{$\checkmark$}&\multicolumn{1}{c|}{$\checkmark$}&1,175&Video \\ \hline
\end{tabular}
\end{table}
\noindent Resources such as the Video Dataset of Incidents (VIDI) \cite{sesver2022vidi}, containing 4,534 clips across 43 incident types, enable extracting meaningful narratives for refined analysis. The ITS series datasets (its4s, its4s2, its4s3) \cite{its_video_quality_data} offer subjective video quality evaluations, critical for assessing the impact of video clarity on user perception. The VCD dataset \cite{devanathan2024seeing}, which gathers e-commerce complaint videos, remains limited due to a lack of detailed complaint descriptions, limiting its utility for less literate users.

These existing studies have integrated modalities such as images, text, and audio to enhance complaint detection via fine-tuning and advanced attention mechanisms. However, there is no existing work and resources, as shown in Table \ref{tab:resource}, to help users write complaints about uploaded videos. To shape the user's perception, we propose a novel task alongside an innovative model with complaint assessment metrics and datasets to advance this field.
\section{\textsc{ComVID} Dataset Details}
\textbf{\textit{Dataset Collection:}} Initially, a dataset comprising 1200 Amazon review videos was assembled, focusing on electronic products such as keyboards, earbuds, mouse, trimmers, and headphones, as well as non-electronic items, including bags, shoes, and household goods. Amazon's extensive reach, particularly in rural regions, provides a valuable platform for analyzing consumer behavior, capturing the essential needs of users such as farmers and laborers for products such as trimmers and shoes. To extract 1- and 2-star reviews, we employed Beautiful Soup for web scraping. The dataset encompasses key attributes, including review ID, rating, review text, aspects, domain name, product name and the associated m3u8 video links. Next, we transcoded these m3u8 URLs into mp4 format for enhanced accessibility. To ensure broad applicability, the dataset includes a balanced distribution of complaint aspects across four domains: Fashion, Electronics, Household, and Others. Key issues span Quality, Functionality, Defects, Missing items, Refunds, and Performance. Fashion items (e.g., shoes, bags) center on quality and design flaws; electronics (e.g., keyboards, trimmers) exhibit a wider range including delays; household products report functional defects; while the Others category captures isolated malfunction cases (e.g., tents). This domain-wise stratification enhances the dataset’s diversity and supports fine-grained complaint analysis.

\begin{figure}[t]    
\centering
    \includegraphics[width=\columnwidth]{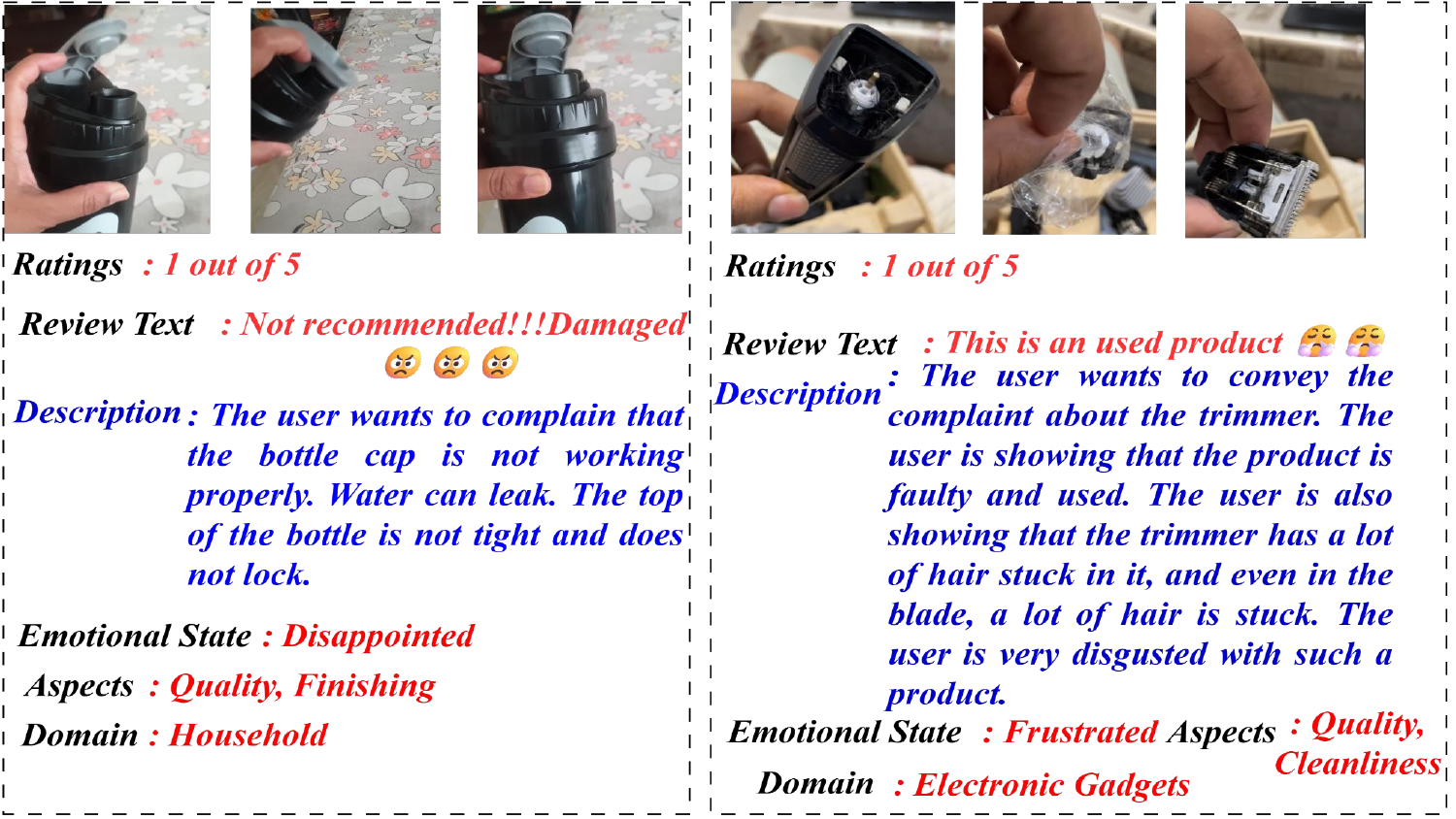}
    \caption{Dataset samples: Video along with corresponding review text, rating, gold label-annotated description, and emotion label. \textit{Note: Emotion, Aspects and Product type labels are assigned after viewing the complete video and reviewing the corresponding review text.}}\label{data_sample}    
\end{figure}
\begin{table}[!b]
\centering
    \caption{Domain wise aspects and product names}
    \label{tab:Ocuurances}
\scriptsize
\begin{tabular}{p{1cm}|p{3cm}|p{2cm}}
\hline
Domain              & Aspects                                                          & Product Name                     \\ \hline
Fashion             & Quality, Functionality, Defective, Design, Missing, Refund            & Shoes, Bag, Watch, T-Shirt          \\ \hline
Electronic Products & Quality, Functionality, Defective, Missing, Performance, Refund, Delay & Mouse, Keyboard, Headphone, Trimmer \\ \hline
House hold          & Functionality, Performance, Defective                              & Bottle, Plates, Plastic Pots       \\ \hline
others              & Defective                                                        & Tent, Rain-Coat                   \\ \hline
\end{tabular}
\end{table}

\noindent After an extensive selection process, 1,175 video samples were incorporated into the final dataset as depicted in Fig.~\ref{data_sample}, each accompanied by annotated descriptions that capture a range of distinct emotional nuances. This selection reflects the high volume and popularity of Amazon reviews, particularly for electronics, where items such as headphones, mouse, and keyboards garner significant attention due to frequent technological advancements. Notably, six samples featured non-English review texts,
further enriching the dataset's linguistic diversity.\\
\textbf{\textit{Dataset Statistics:}} The dataset comprises 655 reviews on electronic gadgets, 273 on household items, 202 on fashion items, and 45 in other categories, offering a comprehensive view of consumer insights across both electronic and non-electronic domains. The focus on electronics, particularly keyboards, mouse, and headphones, highlights their significance in the rapidly growing sector, with evaluations across eight key aspects, including quality, functionality, and defect. These segments represent key areas in an industry undergoing significant growth and technological innovation. 
Table~\ref{tab:Ocuurances} shows domain-wise aspects and product names. By targeting these domains, we aim to assess evolving consumer sentiments and preferences in high-interest areas. 


\noindent\textbf{\textit{Dataset Annotation}}: 
The substantial heterogeneity in complaint structure, user's emotional aspects and linguistic patterns,
necessitates a paradigm of dynamic adaptability in generating complaint descriptions. Tailoring some unyielding approach, such as enforcing a fixed word count \cite{li2017reader}, would fail to account for the inherent linguistic diversity and structural nuances.
A one-size-fits-all strategy can compromise the quality and contextual relevance of generated descriptions, often failing to capture critical details. The collected datasets contain videos but lack adequate textual descriptions. Therefore, our approach prioritizes flexibility in both the length and structure of descriptions, allowing us to effectively convey the unique characteristics, context, and information density inherent to each video. This adaptability ensures that the generated descriptions remain meaningful and capture the essence of the original content. Given these challenges, manual annotation was required to provide accurate text descriptions for the videos.

\textbf{\textit{Phase-1:}} A team of five expert linguists collaborated to generate high-quality complaint descriptions and assign emotional labels. The team included one doctoral-level annotator ({\em Category A}), two annotators with master’s degrees ({\em Category B}), and two undergraduate annotators ({\em Category C}), all proficient in Hindi and English. The {\em Category A} annotator initially provided 50 gold-standard samples as ground truth for reference. Based on these, one {\em Category B} and one {\em Category C} annotator generated descriptions for half of the dataset, while the remaining two annotators rigorously reviewed all descriptions to ensure accuracy, coherence, and adherence to established guidelines. Annotators cross-validated samples they did not annotate, assessing factual accuracy (alignment with video content) and coherence. Only samples with an acceptance tag were included as gold-standard descriptions.

\textbf{\textit{Phase-2:}} Following the description generation, {\em Category A} and {\em B} annotators independently assigned emotional labels, resolving discrepancies through a consensus process. The process achieved a Fleiss' kappa score of 0.64, indicating substantial inter-annotator agreement and ensuring annotation reliability. Annotators validated instances against user-provided content, refined annotations for clarity, and were compensated at \$0.50 per sample, resulting in a robust and unbiased dataset for complaint description and emotion analysis.
\section{Methodology for Generating Complaint Text Description}
\begin{figure*}[hbt]    
\centering
    \includegraphics[scale = 0.20]{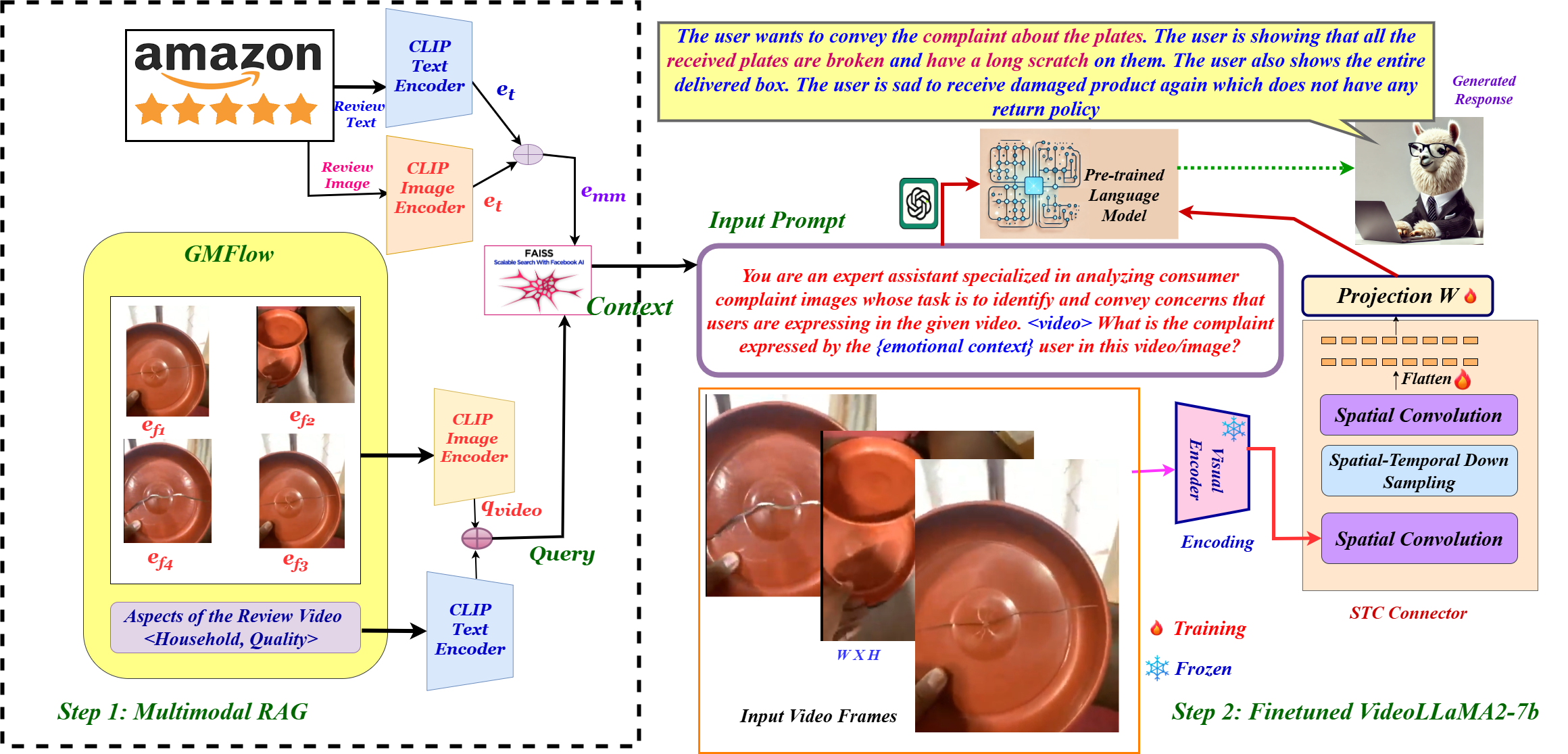}
    \caption{Architectural overview of the multimodal Retrieval-Augmented Generation (RAG) model integrating VideoLLaMA2-7B for generating complaints from user-uploaded videos. The model leverages the Amazon product review dataset containing {review text, image} pairs, which serves as the superset of product reviews included in the \textsc{ComVID} dataset.}\label{architecture}    
\end{figure*}

\textbf{\textit{Problem Statement:}} The video complaint dataset comprises videos \( V \in \mathbb{R}^{F \times 3 \times W \times H} \) and corresponding textual reviews \( R = \{r_1, r_2,\dots, r_m\} \), where \( F \) is the number of sampled frames, $m$ is the number of words and \( W, H \) are the frame width and height, respectively. Each \(R\) is associated with emotion labels $e \in\{$dissatisfaction, blame, frustration, disappointment$\}$, reflecting the user's emotional state. Reviews $R$ are often vague, hence we aim to develop a model \(\mathcal{F}\) that simultaneously processes both the video  \( V \) with textual prompt \( P \) and the associated emotion labels to generate a coherent and contextually accurate descriptive complaint \( Y \), formulated as \( Y = \mathcal{F}(V, P,e) \). The output \( Y \) should accurately encapsulate the user’s complaint by encoding the user's emotional state. 
To address the \textsc{CoD-V} task, the proposed architecture as depicted in Figure \ref{architecture} is designed in the following two steps.
\paragraph{\textbf{Step 1: Developing Multimodal Retrieval Augmented Generation (MR)}:}  
The MR framework is grounded on a large-scale set of publicly accessible Amazon product reviews\footnote{\url{https://cseweb.ucsd.edu/~jmcauley/datasets/amazon_v2}}, utilizing 75.26 million reviews where each user and product has at least five reviews. For complaint analysis, the dataset is further filtered based on three criteria: reviews with a rating of 1, text exceeding 150 characters to ensure detailed complaints, and at least one image per review to capture multimodal cues. The filtering steps ensure the dataset focuses on genuine complaints with adequate contextual and visual detail, resulting in a refined collection of negative reviews, each with comprehensive text and associated images 
\\
\noindent To extract multimodal embeddings, we use  CLIP~\cite{radford2021learning} for both text and image components. The textual content $T$ of a review, consisting of the title $R_{ts}$ and main text $R_t$, is processed by CLIP's pretrained text encoder, yielding the text embedding $\mathbf{e}_t = \mathrm{TextEncoder}(T) \in \mathbb{R}^d$. For image embeddings, each review’s associated images are processed by CLIP’s image encoder to generate $\mathbf{e}_{i_k} = \mathrm{ImageEncoder}(\mathrm{Img}_k)$ $\in \mathbb{R}^d$, and for multiple images their embeddings are averaged: $\bar{\mathbf{e}}_{i} = \frac{1}{n}\sum_{k=1}^n \mathbf{e}_{i_k}$. The final multimodal embedding is obtained by combining the text and image embeddings: $\mathbf{e}_\mathrm{mm} = \frac{1}{2} \left(\mathbf{e}_t + \bar{\mathbf{e}}_i\right)$. Moreover, for efficient similarity search $\mathbf{e}_\mathrm{mm}$ are further aggregated into matrix $E$ and indexed using FAISS~\cite{johnson2019billion} defined as $\mathrm{Index} = \mathrm{FAISSIndexFlatL2}(E)$.
 
We encode the review videos by employing GMFlow~\cite{xu2022gmflow} and extract four keyframes, further process them with CLIP’s image encoder to obtain embeddings $\mathbf{e}_{f_1}, \mathbf{e}_{f_2}, \mathbf{e}_{f_3}, \mathbf{e}_{f_4}$, followed by aggregating them into $\mathbf{q}_\mathrm{video} = \frac{1}{4} \sum_{i=1}^4 (\mathbf{e}_{f_i})$. 
Subsequently, we take corresponding product aspects and pass through CLIP’s text encoder to produce $\mathbf{e}_\mathrm{textual}$, resulting into the final query embedding as: $\mathbf{q} = \alpha \,\mathbf{q}_\mathrm{video} + (1-\alpha) \,\mathbf{e}_\mathrm{textual}$. The top $k$ nearest neighbors $\{\mathbf{e}_\mathrm{mm}^{(i_1)}, \dots, \mathbf{e}_\mathrm{mm}^{(i_k)}\}$ are retrieved from the FAISS index, providing retrieval-augmented context for the query.
\paragraph{\textbf{Step 2: Supervised Fine-Tuning and Generation}:} VideoLLaMA2-7b is fine-tuned on supervised video-text pairs. During inference, the query prompt is enhanced with the user's emotional state (e.g., frustration) and the top-$k$ retrieved complaint reviews from MR. The model processes this enriched input to generate a refined output, effectively integrating textual, visual, and domain-specific context. Fig. \ref{architecture} illustrates the overall framework.

\section{Experimental Results and Analysis}
We describe our experimental setup, comparative baselines, empirical results, and performance analysis of fine-tuned VideoLLaMA2-7b+MR for the novel \textsc{CoD-V} task. We conduct qualitative evaluations to scrutinize result quality and discuss observed performances. Our research addresses the following questions:\\
\textbf{RQ1:} Can the fine-tuned VideoLLaMA2-7b+MR surpass conventional baselines and state-of-the-art methods in technical performance, serving as a superior method for the \textsc{CoD-V} task?\\
\textbf{RQ2:} Can providing emotion as extra input help? How do various models perform based on human perception?\\
\textbf{RQ3:} How does \textit{CoD-V} task differ from traditional summary generation and video description?\\
\textbf{RQ4:} What are the societal benefits of the proposed work, and how extensive is the model's applicability across diverse NLP tasks? \\ 
We use the following multimodal backbone models: fine-tuned VideoLLaVA2-7b \cite{zhu2023languagebind}, the fine-tuned BLIP-VQA-BASE \cite{li2023blip2}, Qwen2-VL-7b \cite{Qwen2VL}, Gemma3-12b \cite{gemma_2025} and the integration of LSTM networks with VGG16 \cite{simonyan2014very} and ResNet50 \cite{he2016deep}.
\subsection{Evaluation Metrics}
We propose a standard evaluation metric to assess the nature of the complaint retained (CR) in the generated output, focusing on sentiment score, emotion detection, and aspect identification, compared against the gold label aspects.  

\noindent\textbf{Sentiment score}: To analyze the sentiment, we utilize Vader score~\cite{hutto2014vader}. We first calculate Vader score \( S_{\text{Vader},i} \) for a sentence in the generated inferences. This score ranges from -1 to 1, so we normalize it to a range from 0 to 1 as: $
S_{\text{N Vader}, i} = \frac{S_{\text{Vader}, i} + 1}{2}$. Next, the average normalized sentiment score for each $k$-th sample with $N_k$ sentences is calculated as $S_{\text{N Vader, $k$}} = \frac{1}{N_k} \sum_{i=1}^{N_k} S_{\text{N Vader}, i}$. Finally, we compute the overall average Vader sentiment score $S_{\text{N Vader}}$ as the average across all inference samples.

\noindent\textbf{Emotional intensity} is a critical component of complaints. We analyze predicted text using \texttt{Text2Emotion} library, which extracts scores for four key emotions: Happy (H), Angry (A), Surprise (S), and Fear (F). Each emotion score lies in $[0, 1]$, and Emotion Score (ES) for a sample $k$ is computed as the average of all emotion scores: 
$ES_k = \frac{H_k + A_k + S_k + F_k}{4}$. This provides an aggregate measure of emotional intensity in the generated complaint.

\noindent\textbf{Aspect Score (AS)} measures how effectively a generated complaint retains key aspects from the original complaint. Since complaints often focus on specific issues (e.g., product quality, product defects), it is crucial to ensure that these ground-truth aspects are preserved in the predicted text.
To evaluate this, we use GPT-4 as a classifier. For each manually annotated ground-truth aspect, we provide the following prompt: ``Is the aspect `{predicted\_aspect}' present in the given `{text}'? Answer with only `Yes' or `No'.'' GPT-4 responds with either Yes or No, which is mapped to a binary score of 1 or 0 respectively. The final Aspect Score (AS) for the $k$-th sample is computed as the fraction of ground-truth aspects that are retained in the predicted text:
$AS_k = \frac{\sum_{i=1}^{N_k} \mathbf{1}(\text{Aspect}_i \in T_p)}{N_k}$
where \( N_k \) is the total number of ground-truth aspects, and \( \mathbf{1}(\cdot) \) is an indicator function. 
Aspect score provides a structured method for assessing how well the generated complaint preserves essential details compared to the original. 

\noindent\textbf{Complaint retention (CR) score} is calculated as $\text{CR score} = \frac{S_{\text{N Vader}} + ES + AS}{3}$. Thus, the complaint retention score combines the three metrics (normalized sentiment, emotion, and aspect presence) into a single metric, and gives a balanced evaluation of the nature of the complaint retained.

\noindent\textbf{Standard NLG metrics}: Further, we evaluate model performance using 8 standard metrics: BLEU \cite{lin-och-2004-orange}, ROUGE \cite{lin-2004-rouge}, BERTScore \cite{zhang2019bertscore}, MoverScore \cite{zhao2019moverscore} METEOR \cite{banerjee2005meteor}, Perplexity \cite{jelinek1977perplexity}, Flesch Reading Ease \cite{farr1951simplification}, and Coleman-Liau Index \cite{coleman1975computer}. BLEU and ROUGE measure n-gram similarity, BERTScore (BS) and MoverScore (MoS) evaluate semantic alignment via contextual embeddings, METEOR (MeS) captures precision-recall with synonym flexibility, Perplexity (PS) quantifies fluency, while Flesch Reading Ease (FRES) and Coleman-Liau (CLIS) assess the readability of generated content.\\
\textbf{\textit{Experimental Configuration:}} To ensure a fair comparison among all models, both VideoLLAMA2-7b and its LoRA fine-tuned variant were trained with these hyperparameters: LoRA rank set to 64, LoRA alpha at 128, a batch size of 16, and a learning rate (LR) of \(2 \times 10^{-4}\) with a cosine learning rate scheduler. We use LR of \(2 \times 10^{-5}\) for multi-modal projector layer. We trained for 10 epochs.
\begin{table*}
\caption{Main results: Proposed framework vs popular fine-tuned visual language models; (R1-RL): ROUGE variants; (B1-B3): BLEU variants; BS: BERTScore; FRES: Flesch Reading Ease Score, CLIS: Coleman-Liau Index Score; PS: Perplexity Score, MoS:  Mover Score and MeS: Meteor Score. Best scores are highlighted in bold. Second best are underlined.}\label{tab:ablation}
\centering

\scriptsize
\begin{tabular}{l|l|rrrrrrrrrrrr}
\hline
\textbf{Setting}& \textbf{Model Name}& \multicolumn{1}{l|}{\textbf{R1}$\uparrow$} & \multicolumn{1}{l|}{\textbf{R2}$\uparrow$} & \multicolumn{1}{l|}{\textbf{RL}$\uparrow$} & \multicolumn{1}{l|}{\textbf{B1}$\uparrow$} & \multicolumn{1}{l|}{\textbf{B2}$\uparrow$} & \multicolumn{1}{l|}{\textbf{BL}$\uparrow$} & \multicolumn{1}{l|}{\textbf{BS}$\uparrow$} & \multicolumn{1}{l|}{\textbf{FRES}$\uparrow$} & \multicolumn{1}{l|}{\textbf{CLRS}$\downarrow$} & \multicolumn{1}{l|}{\textbf{PS}$\downarrow$} & \multicolumn{1}{l|}{\textbf{MeS}$\uparrow$} & \multicolumn{1}{l}{\textbf{MoS}$\uparrow$} \\ \hline 
\multirow{12}{*}{No Emotion}       & LSTM+VGG16 & \multicolumn{1}{r|}{0.41} & \multicolumn{1}{r|}{0.3}  & \multicolumn{1}{r|}{0.38} & \multicolumn{1}{r|}{0.62} & \multicolumn{1}{r|}{0.53} & \multicolumn{1}{r|}{0.48} & \multicolumn{1}{r|}{0.72} & \multicolumn{1}{r|}{66.47} & \multicolumn{1}{r|}{\underline{5.42}}    & \multicolumn{1}{r|}{86.36}    & \multicolumn{1}{r|}{0.42} & -0.39     \\ \cline{2-14} 
  & LSTM+ResNet50       & \multicolumn{1}{r|}{0.42} & \multicolumn{1}{r|}{0.32} & \multicolumn{1}{r|}{0.39} & \multicolumn{1}{r|}{0.64} & \multicolumn{1}{r|}{0.55} & \multicolumn{1}{r|}{0.48} & \multicolumn{1}{r|}{0.72} & \multicolumn{1}{r|}{65.84} & \multicolumn{1}{r|}{5.57} & \multicolumn{1}{r|}{87.95}    & \multicolumn{1}{r|}{0.42} & -0.37     \\ \cline{2-14} 
  & BLIP-VQA-BASE& \multicolumn{1}{r|}{0.5}  & \multicolumn{1}{r|}{0.34} & \multicolumn{1}{r|}{0.45} & \multicolumn{1}{r|}{0.6}  & \multicolumn{1}{r|}{0.53} & \multicolumn{1}{r|}{0.47} & \multicolumn{1}{r|}{0.91} & \multicolumn{1}{r|}{\underline{81.87}} & \multicolumn{1}{r|}{5.95} & \multicolumn{1}{r|}{80.28}    & \multicolumn{1}{r|}{0.39} & 0.07      \\ \cline{2-14} 
  & QWEN2-VL-7b& \multicolumn{1}{r|}{0.45} & \multicolumn{1}{r|}{0.32} & \multicolumn{1}{r|}{0.42} & \multicolumn{1}{r|}{0.61} & \multicolumn{1}{r|}{0.52} & \multicolumn{1}{r|}{0.48} & \multicolumn{1}{r|}{0.90} & \multicolumn{1}{r|}{70.48} & \multicolumn{1}{r|}{8.91} & \multicolumn{1}{r|}{{\textbf{70.12}}} & \multicolumn{1}{r|}{0.30} & 0.05      \\ \cline{2-14} 
  & Gemma3-12b & \multicolumn{1}{r|}{0.46} & \multicolumn{1}{r|}{0.33} & \multicolumn{1}{r|}{0.45} & \multicolumn{1}{r|}{0.61} & \multicolumn{1}{r|}{0.52} & \multicolumn{1}{r|}{0.51} & \multicolumn{1}{r|}{0.90} & \multicolumn{1}{r|}{73.45} & \multicolumn{1}{r|}{9.11} & \multicolumn{1}{r|}{78.44}    & \multicolumn{1}{r|}{0.41} & 0.08      \\ \cline{2-14} 
  & VideoLLaVA2-7b      & \multicolumn{1}{r|}{0.5}  & \multicolumn{1}{r|}{0.38} & \multicolumn{1}{r|}{0.49} & \multicolumn{1}{r|}{0.65} & \multicolumn{1}{r|}{0.54} & \multicolumn{1}{r|}{0.53} & \multicolumn{1}{r|}{0.91} & \multicolumn{1}{r|}{76.9}  & \multicolumn{1}{r|}{7.55} & \multicolumn{1}{r|}{120.4}    & \multicolumn{1}{r|}{0.44} & 0.115     \\ \cline{2-14} 
  & VideoLLaMA2-7b      & \multicolumn{1}{r|}{0.52} & \multicolumn{1}{r|}{0.38} & \multicolumn{1}{r|}{0.50}  & \multicolumn{1}{r|}{0.64} & \multicolumn{1}{r|}{0.56} & \multicolumn{1}{r|}{0.53} & \multicolumn{1}{r|}{0.90} & \multicolumn{1}{r|}{77.01} & \multicolumn{1}{r|}{7.41} & \multicolumn{1}{r|}{96.29}    & \multicolumn{1}{r|}{0.43} & 0.12      \\ \cline{2-14} 
  & BLIP-VQA-BASE+ MR        & \multicolumn{1}{r|}{0.52} & \multicolumn{1}{r|}{0.36} & \multicolumn{1}{r|}{0.47} & \multicolumn{1}{r|}{0.63} & \multicolumn{1}{r|}{0.55} & \multicolumn{1}{r|}{0.49} & \multicolumn{1}{r|}{0.91} & \multicolumn{1}{r|}{\textbf{82.11}}    & \multicolumn{1}{r|}{7.98} & \multicolumn{1}{r|}{82.76}    & \multicolumn{1}{r|}{0.41} & 0.09      \\ \cline{2-14} 
  & QWEN2-VL-7b+MR      & \multicolumn{1}{r|}{0.47} & \multicolumn{1}{r|}{0.35} & \multicolumn{1}{r|}{0.43} & \multicolumn{1}{r|}{0.63} & \multicolumn{1}{r|}{0.54} & \multicolumn{1}{r|}{0.5}  & \multicolumn{1}{r|}{0.90} & \multicolumn{1}{r|}{72.62} & \multicolumn{1}{r|}{10.11}& \multicolumn{1}{r|}{72.55}    & \multicolumn{1}{r|}{0.33} & 0.08      \\ \cline{2-14} 
  & Gemma3-12b+MR       & \multicolumn{1}{r|}{0.48} & \multicolumn{1}{r|}{0.39} & \multicolumn{1}{r|}{0.47} & \multicolumn{1}{r|}{0.63} & \multicolumn{1}{r|}{0.54} & \multicolumn{1}{r|}{0.52} & \multicolumn{1}{r|}{0.89} & \multicolumn{1}{r|}{79.21} & \multicolumn{1}{r|}{9.14} & \multicolumn{1}{r|}{80.44}    & \multicolumn{1}{r|}{0.43} & 0.09      \\ \cline{2-14} 
  & VideoLLaVA2-7b+MR   & \multicolumn{1}{r|}{0.53} & \multicolumn{1}{r|}{0.39} & \multicolumn{1}{r|}{0.49} & \multicolumn{1}{r|}{0.63} & \multicolumn{1}{r|}{0.55} & \multicolumn{1}{r|}{\underline{0.54}}    & \multicolumn{1}{r|}{\underline{0.92}}    & \multicolumn{1}{r|}{78.11} & \multicolumn{1}{r|}{7.09} & \multicolumn{1}{r|}{121.14}   & \multicolumn{1}{r|}{0.44} & 0.14      \\ \cline{2-14} 
 & \textbf{\begin{tabular}[c]{@{}l@{}}VideoLLaMA2-7b+MR\\ (Proposed)\end{tabular}}        & \multicolumn{1}{r|}{\underline{0.55}}    & \multicolumn{1}{r|}{\underline{0.40}}    & \multicolumn{1}{r|}{\underline{0.51}}    & \multicolumn{1}{r|}{\underline{0.65}}    & \multicolumn{1}{r|}{\underline{0.57}}    & \multicolumn{1}{r|}{0.53} & \multicolumn{1}{r|}{0.91} & \multicolumn{1}{r|}{78.07} & \multicolumn{1}{r|}{7.26} & \multicolumn{1}{r|}{95.39}    & \multicolumn{1}{r|}{0.46}    & {0.14}\\ \hline
\multirow{12}{*}{\textbf{Emotion}} & LSTM+VGG16 & \multicolumn{1}{r|}{0.43} & \multicolumn{1}{r|}{0.31} & \multicolumn{1}{r|}{0.4}  & \multicolumn{1}{r|}{0.63} & \multicolumn{1}{r|}{0.54} & \multicolumn{1}{r|}{0.49} & \multicolumn{1}{r|}{0.73} & \multicolumn{1}{r|}{67.72} & \multicolumn{1}{r|}{\textbf{5.32}} & \multicolumn{1}{r|}{88.14}    & \multicolumn{1}{r|}{0.43} & -0.37     \\ \cline{2-14} 
  & LSTM+ResNet50       & \multicolumn{1}{r|}{0.43} & \multicolumn{1}{r|}{0.33} & \multicolumn{1}{r|}{0.42} & \multicolumn{1}{r|}{0.65} & \multicolumn{1}{r|}{0.56} & \multicolumn{1}{r|}{0.5}  & \multicolumn{1}{r|}{0.74} & \multicolumn{1}{r|}{69.35} & \multicolumn{1}{r|}{5.62} & \multicolumn{1}{r|}{89.03}    & \multicolumn{1}{r|}{0.44} & -0.36     \\ \cline{2-14} 
  & BLIP-VQA-BASE& \multicolumn{1}{r|}{0.51} & \multicolumn{1}{r|}{0.36} & \multicolumn{1}{r|}{0.47} & \multicolumn{1}{r|}{0.62} & \multicolumn{1}{r|}{0.55} & \multicolumn{1}{r|}{0.5}  & \multicolumn{1}{r|}{0.91} & \multicolumn{1}{r|}{78.83} & \multicolumn{1}{r|}{6.8}  & \multicolumn{1}{r|}{118.56}   & \multicolumn{1}{r|}{0.4}  & 0.11      \\ \cline{2-14} 
  & QWEN2-VL-7b& \multicolumn{1}{r|}{0.48} & \multicolumn{1}{r|}{0.37} & \multicolumn{1}{r|}{0.5}  & \multicolumn{1}{r|}{0.6}  & \multicolumn{1}{r|}{0.54} & \multicolumn{1}{r|}{0.49} & \multicolumn{1}{r|}{0.91} & \multicolumn{1}{r|}{71.38} & \multicolumn{1}{r|}{8.93} & \multicolumn{1}{r|}{\underline{71.23}}    & \multicolumn{1}{r|}{0.31} & 0.06      \\ \cline{2-14} 
  & Gemma3-12b & \multicolumn{1}{r|}{0.49} & \multicolumn{1}{r|}{0.4}  & \multicolumn{1}{r|}{0.48} & \multicolumn{1}{r|}{0.64} & \multicolumn{1}{r|}{0.55} & \multicolumn{1}{r|}{0.53} & \multicolumn{1}{r|}{0.91} & \multicolumn{1}{r|}{76.22} & \multicolumn{1}{r|}{9.15} & \multicolumn{1}{r|}{80.45}    & \multicolumn{1}{r|}{0.44} & 0.1       \\ \cline{2-14} 
  & VideoLLaVA2-7b      & \multicolumn{1}{r|}{0.54} & \multicolumn{1}{r|}{0.42} & \multicolumn{1}{r|}{0.53} & \multicolumn{1}{r|}{0.64} & \multicolumn{1}{r|}{0.55} & \multicolumn{1}{r|}{0.57} & \multicolumn{1}{r|}{0.92} & \multicolumn{1}{r|}{77.15} & \multicolumn{1}{r|}{7.29} & \multicolumn{1}{r|}{116.2}    & \multicolumn{1}{r|}{0.44} & 0.16      \\ \cline{2-14} 
  & VideoLLaMA2-7b      & \multicolumn{1}{r|}{0.56} & \multicolumn{1}{r|}{0.42} & \multicolumn{1}{r|}{0.49} & \multicolumn{1}{r|}{0.64} & \multicolumn{1}{r|}{0.56} & \multicolumn{1}{r|}{0.57} & \multicolumn{1}{r|}{0.92} & \multicolumn{1}{r|}{79.15} & \multicolumn{1}{r|}{7.01} & \multicolumn{1}{r|}{111.1}    & \multicolumn{1}{r|}{\underline{0.47}} & 0.19      \\ \cline{2-14} 
  & BLIP-VQA-BASE+ MR        & \multicolumn{1}{r|}{0.53} & \multicolumn{1}{r|}{0.38} & \multicolumn{1}{r|}{0.49} & \multicolumn{1}{r|}{0.64} & \multicolumn{1}{r|}{0.57} & \multicolumn{1}{r|}{0.52} & \multicolumn{1}{r|}{0.91} & \multicolumn{1}{r|}{77.85} & \multicolumn{1}{r|}{6.82} & \multicolumn{1}{r|}{118.58}   & \multicolumn{1}{r|}{0.42} & 0.13      \\ \cline{2-14} 
  & QWEN2-VL-7b+MR      & \multicolumn{1}{r|}{0.50} & \multicolumn{1}{r|}{0.39} & \multicolumn{1}{r|}{0.52} & \multicolumn{1}{r|}{0.62} & \multicolumn{1}{r|}{0.56} & \multicolumn{1}{r|}{0.51} & \multicolumn{1}{r|}{0.91} & \multicolumn{1}{r|}{71.4}  & \multicolumn{1}{r|}{8.95} & \multicolumn{1}{r|}{71.25}    & \multicolumn{1}{r|}{0.33} & 0.08      \\ \cline{2-14} 
  & Gemma3-12b+MR       & \multicolumn{1}{r|}{0.51} & \multicolumn{1}{r|}{0.42} & \multicolumn{1}{r|}{0.50} & \multicolumn{1}{r|}{0.66} & \multicolumn{1}{r|}{0.57} & \multicolumn{1}{r|}{0.55} & \multicolumn{1}{r|}{0.92} & \multicolumn{1}{r|}{79.24} & \multicolumn{1}{r|}{9.17} & \multicolumn{1}{r|}{80.47}    & \multicolumn{1}{r|}{0.46} & 0.12      \\ \cline{2-14} 
  & VideoLLaVA2-7b+MR   & \multicolumn{1}{r|}{0.57} & \multicolumn{1}{r|}{0.46} & \multicolumn{1}{r|}{0.54} & \multicolumn{1}{r|}{0.68} & \multicolumn{1}{r|}{0.61} & \multicolumn{1}{r|}{0.58} & \multicolumn{1}{r|}{0.92} & \multicolumn{1}{r|}{78.14} & \multicolumn{1}{r|}{7.35} & \multicolumn{1}{r|}{114.51}   & \multicolumn{1}{r|}{\textbf{0.51}} & \underline{0.21}      \\ \cline{2-14} 
  & \textbf{\begin{tabular}[c]{@{}l@{}}VideoLLaMA2-7b+MR\\ (Proposed)\end{tabular}} & \multicolumn{1}{r|}{\textbf{0.59}} & \multicolumn{1}{r|}{\textbf{0.47}} & \multicolumn{1}{r|}{\textbf{0.56}} & \multicolumn{1}{r|}{\textbf{0.69}} & \multicolumn{1}{r|}{\textbf{0.63}} & \multicolumn{1}{r|}{\textbf{0.59}} & \multicolumn{1}{r|}{\textbf{0.93}} & \multicolumn{1}{r|}{79.58} & \multicolumn{1}{r|}{7.27} & \multicolumn{1}{r|}{97.16}    & \multicolumn{1}{r|}{\textbf{0.51}} & \textbf{0.24}\\ \hline
\end{tabular}
\end{table*}

\begin{table*}[t]
\caption{Comparison across tasks: Proposed \textit{CoD-V} vs Summary Generation (SG) and Video Description (VD) on popular visual language models in zero-shot setting; VS: Vader Score, CR: Complaint Retention. Best scores are highlighted in bold.}\label{task_difference}
\centering
\scriptsize
\tabcolsep5pt
\begin{tabular}{l|l|rrrrrrrrrrrrrr}
\hline
\textbf{Model Name } &\textbf{Task}  & \multicolumn{1}{l|}{\textbf{R1}$\uparrow$} & \multicolumn{1}{l|}{\textbf{R2}$\uparrow$} & \multicolumn{1}{l|}{\textbf{RL}$\uparrow$} & \multicolumn{1}{l|}{\textbf{B1}$\uparrow$} & \multicolumn{1}{l|}{\textbf{B2}$\uparrow$} & \multicolumn{1}{l|}{\textbf{BL}$\uparrow$} & \multicolumn{1}{l|}{\textbf{BS}$\uparrow$} & \multicolumn{1}{l|}{\textbf{FRES}$\uparrow$} & \multicolumn{1}{l|}{\textbf{CLRS}$\downarrow$} & \multicolumn{1}{l|}{\textbf{PS}$\downarrow$} & \multicolumn{1}{l|}{\textbf{MeS}$\uparrow$} & \multicolumn{1}{l}{\textbf{MoS}$\uparrow$} & \multicolumn{1}{l|}{\textbf{VS}$\uparrow$}  & \multicolumn{1}{l}{\textbf{CR}$\uparrow$} \\ \hline
{VideoLLaVA2-7b} & SG  & \multicolumn{1}{r|}{0.21} & \multicolumn{1}{r|}{0.02} & \multicolumn{1}{r|}{0.17} & \multicolumn{1}{r|}{0.55} & \multicolumn{1}{r|}{0.43} & \multicolumn{1}{r|}{0.32} & \multicolumn{1}{r|}{0.85} & \multicolumn{1}{r|}{68.51} & \multicolumn{1}{r|}{7.58} & \multicolumn{1}{r|}{\textbf{15.05}} & \multicolumn{1}{r|}{0.14} & \multicolumn{1}{r|}{-0.1}& \multicolumn{1}{r|}{0.06}& 0.41  \\ 
  & VD  & \multicolumn{1}{r|}{0.15} & \multicolumn{1}{r|}{0.02} & \multicolumn{1}{r|}{0.12} & \multicolumn{1}{r|}{0.27} & \multicolumn{1}{r|}{0.23} & \multicolumn{1}{r|}{0.19} & \multicolumn{1}{r|}{0.84} & \multicolumn{1}{r|}{61.10}  & \multicolumn{1}{r|}{8.60}  & \multicolumn{1}{r|}{34.40}  & \multicolumn{1}{r|}{0.15} & \multicolumn{1}{r|}{-0.13}  & \multicolumn{1}{r|}{0.16}  & 0.48  \\
  & \textsc{CoD-V}  & \multicolumn{1}{r|}{0.31} & \multicolumn{1}{r|}{0.01} & \multicolumn{1}{r|}{0.24} & \multicolumn{1}{r|}{0.50}  & \multicolumn{1}{r|}{0.42} & \multicolumn{1}{r|}{\textbf{0.39}} & \multicolumn{1}{r|}{\textbf{0.89}} & \multicolumn{1}{r|}{57.80}  & \multicolumn{1}{r|}{10.80} & \multicolumn{1}{r|}{34.30}  & \multicolumn{1}{r|}{0.16} & \multicolumn{1}{r|}{\textbf{0.00}}& \multicolumn{1}{r|}{-0.14}  & 0.57  \\ \hline
{GPT4O} & SG  & \multicolumn{1}{r|}{0.21} & \multicolumn{1}{r|}{0.02} & \multicolumn{1}{r|}{0.15} & \multicolumn{1}{r|}{0.54} & \multicolumn{1}{r|}{0.44} & \multicolumn{1}{r|}{0.33} & \multicolumn{1}{r|}{0.84} & \multicolumn{1}{r|}{55.60}  & \multicolumn{1}{r|}{10.30} & \multicolumn{1}{r|}{41.20}  & \multicolumn{1}{r|}{0.14} & \multicolumn{1}{r|}{-0.10}& \multicolumn{1}{r|}{0.30} & 0.45  \\ 
  & VD  & \multicolumn{1}{r|}{0.14} & \multicolumn{1}{r|}{0.02} & \multicolumn{1}{r|}{0.10}  & \multicolumn{1}{r|}{0.42} & \multicolumn{1}{r|}{0.33} & \multicolumn{1}{r|}{0.23} & \multicolumn{1}{r|}{0.83} & \multicolumn{1}{r|}{52.80}  & \multicolumn{1}{r|}{9.60}  & \multicolumn{1}{r|}{31.04} & \multicolumn{1}{r|}{0.12} & \multicolumn{1}{r|}{-0.13}  & \multicolumn{1}{r|}{0.22}  & 0.42  \\ 
  & \textsc{CoD-V}  & \multicolumn{1}{r|}{0.28} & \multicolumn{1}{r|}{0.06} & \multicolumn{1}{r|}{0.19} & \multicolumn{1}{r|}{\textbf{0.59}} & \multicolumn{1}{r|}{\textbf{0.49}} & \multicolumn{1}{r|}{\textbf{0.39}} & \multicolumn{1}{r|}{0.86} & \multicolumn{1}{r|}{51.50}  & \multicolumn{1}{r|}{10.80} & \multicolumn{1}{r|}{38.04} & \multicolumn{1}{r|}{0.16} & \multicolumn{1}{r|}{-0.04}  & \multicolumn{1}{r|}{-0.26}  & 0.56  \\ \hline
{GeminiFlash} & SG  & \multicolumn{1}{r|}{0.22} & \multicolumn{1}{r|}{0.03} & \multicolumn{1}{r|}{0.17} & \multicolumn{1}{r|}{0.49} & \multicolumn{1}{r|}{0.40}  & \multicolumn{1}{r|}{0.31} & \multicolumn{1}{r|}{0.84} & \multicolumn{1}{r|}{54.60}  & \multicolumn{1}{r|}{10.50} & \multicolumn{1}{r|}{39.11} & \multicolumn{1}{r|}{0.16} & \multicolumn{1}{r|}{-0.08}  & \multicolumn{1}{r|}{-0.01} & 0.47  \\ 
  & VD  & \multicolumn{1}{r|}{0.22} & \multicolumn{1}{r|}{0.03} & \multicolumn{1}{r|}{0.16} & \multicolumn{1}{r|}{0.50}  & \multicolumn{1}{r|}{0.41} & \multicolumn{1}{r|}{0.31} & \multicolumn{1}{r|}{0.85} & \multicolumn{1}{r|}{54.30}  & \multicolumn{1}{r|}{10.60} & \multicolumn{1}{r|}{42.40}  & \multicolumn{1}{r|}{0.15} & \multicolumn{1}{r|}{-0.08}  & \multicolumn{1}{r|}{-0.01}  & 0.48  \\ 
  & \textsc{CoD-V}  & \multicolumn{1}{r|}{0.28} & \multicolumn{1}{r|}{0.07} & \multicolumn{1}{r|}{0.19} & \multicolumn{1}{r|}{0.53} & \multicolumn{1}{r|}{0.45} & \multicolumn{1}{r|}{0.36} & \multicolumn{1}{r|}{0.86} & \multicolumn{1}{r|}{55.50}  & \multicolumn{1}{r|}{10.60} & \multicolumn{1}{r|}{42.50}  & \multicolumn{1}{r|}{0.20}  & \multicolumn{1}{r|}{\textbf{0.00}} & \multicolumn{1}{r|}{-0.24}  & 0.55  \\ \hline
{\textbf{Proposed}} & SG  & \multicolumn{1}{r|}{0.32} & \multicolumn{1}{r|}{0.11} & \multicolumn{1}{r|}{0.26} & \multicolumn{1}{r|}{0.53} & \multicolumn{1}{r|}{0.44} & \multicolumn{1}{r|}{0.36} & \multicolumn{1}{r|}{0.88} & \multicolumn{1}{r|}{78.40}  & \multicolumn{1}{r|}{5.80}  & \multicolumn{1}{r|}{42.20}  & \multicolumn{1}{r|}{0.24} & \multicolumn{1}{r|}{-0.01} & \multicolumn{1}{r|}{0.05}& 0.59  \\ 
  & VD  & \multicolumn{1}{r|}{0.25} & \multicolumn{1}{r|}{0.08} & \multicolumn{1}{r|}{0.20}  & \multicolumn{1}{r|}{0.39} & \multicolumn{1}{r|}{0.34} & \multicolumn{1}{r|}{0.28} & \multicolumn{1}{r|}{0.86} & \multicolumn{1}{r|}{70.10}  & \multicolumn{1}{r|}{7.40}  & \multicolumn{1}{r|}{41.10}  & \multicolumn{1}{r|}{0.22} & \multicolumn{1}{r|}{-0.04} & \multicolumn{1}{r|}{0.14}& 0.58  \\ 
  & \textsc{CoD-V}  & \multicolumn{1}{r|}{\textbf{0.39}} & \multicolumn{1}{r|}{\textbf{0.27}} & \multicolumn{1}{r|}{\textbf{0.36}} & \multicolumn{1}{r|}{0.49} & \multicolumn{1}{r|}{0.43} & \multicolumn{1}{r|}{\textbf{0.39}} & \multicolumn{1}{r|}{\textbf{0.89}} & \multicolumn{1}{r|}{\textbf{79.91}}  & \multicolumn{1}{r|}{\textbf{5.50}}  & \multicolumn{1}{r|}{41.60}  & \multicolumn{1}{r|}{\textbf{0.34}} & \multicolumn{1}{r|}{-0.01}  & \multicolumn{1}{r|}{\textbf{-0.28}}  & \textbf{0.62}  \\ \hline
\end{tabular}
\end{table*}

\begin{table*}[!t]
\centering
\scriptsize
\caption{Qualitative Analysis for a sample. The section highlighted in blue indicates the observed performance improvements. Ground truth: ``The user wants to convey about the complaint of mouse.  {\color{blue}{The user}} {\color{blue}{claims that the}} {\color{blue}{mouse  is defective.}} He has received a damaged product. The scrolling  button is not working properly.''}\label{qualitative}
\begin{tabular}
{l|l|l|l|l}
\hline
\textbf{Video} & &\multicolumn{1}{l|}{BLIP-VQA-BASE} & \multicolumn{1}{l|}{VideoLLaVA2-7b}& VideoLLaMA2-7b+MR \\ \hline
\multirow{12}{*}
{\includegraphics[scale = 0.15]{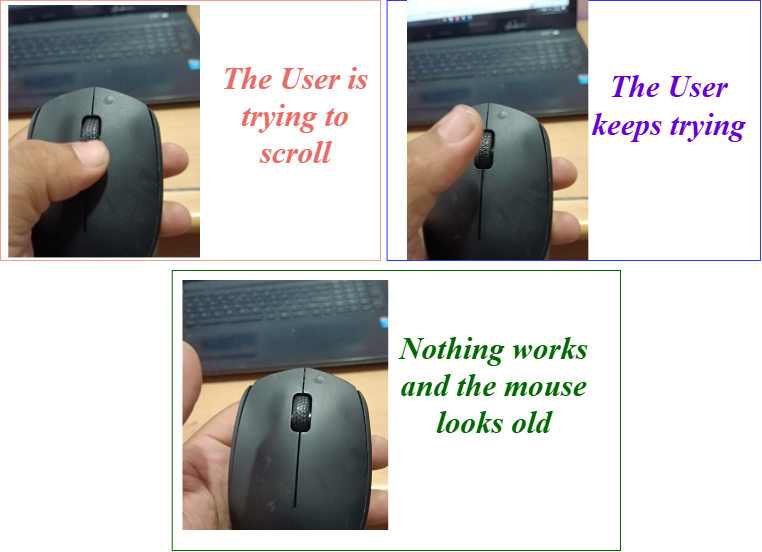}}                                    & \multirow{5}{*}{\rotatebox{90}{Without Emotion}} &the user wants to convey about & The user wants to convey about & The user wants to convey   about\\ &&  the complaint of mouse. the & the complaint of  mouse is old. & the complaint of mouse.   The user \\&& user is saying that the mouse &    The user has received an old product. &   shows the quality of   the mouse \\&& scroller is not working  and the &  Which is not having good quality. & which is not good   by placing \\ && scroll button is not working.&  The scrolling button is not working &  it in his hand.  The sensor is \\ 
&&&as well.&not working\\
\cline{2-5}
&\multirow{6}{*}{\rotatebox{90}{With Emotion}}& the user wants to convey about & The user wants to convey  about & The user wants to convey  about \\
&&the complaint of  mouse. the  &the complaint of mouse. The user& the complaint of mouse.{\color{blue}{The user}}\\
&& user shows the quality of the & has   dissatisfaction regarding the&{\color{blue}{claims that the mouse is defective}}\\
&&mouse which is not good.the& product durability.&{\color{blue}{and he has received a damaged}}\\
&& user has shown this by taking&&{\color{blue}{product}}. The scrolling button is\\
&&the mouse in his hand.&&not working properly.\\ \hline
\end{tabular}
\end{table*}

\begin{figure*}[]    
\centering
    \includegraphics[width=0.5\textwidth]{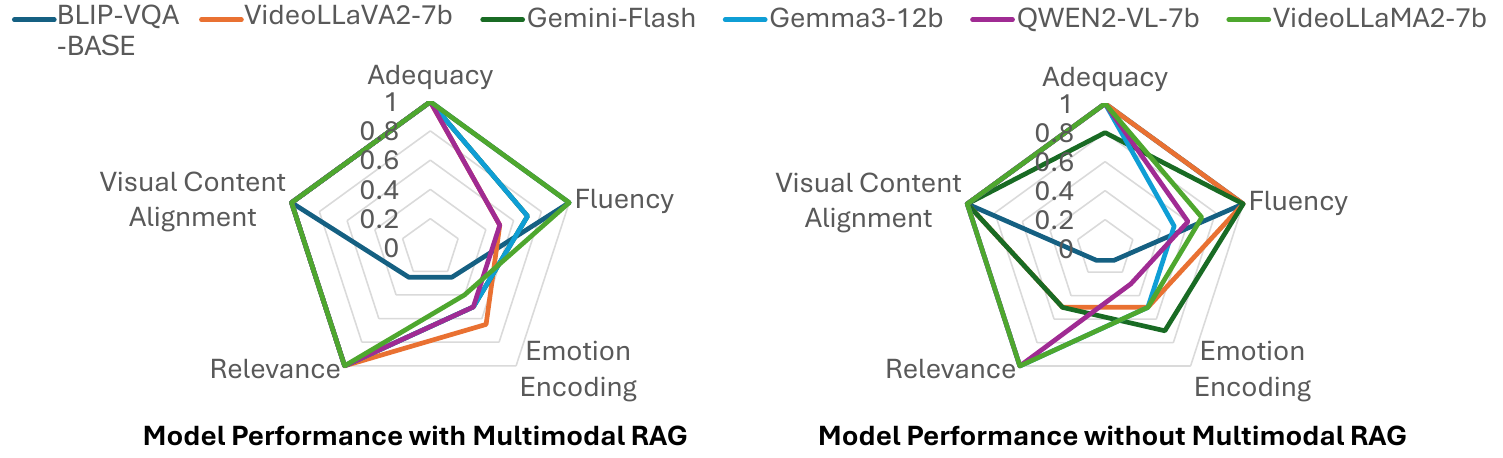}
    \caption{Illustration of Human Evaluation among popular VLM vs the proposed model. }\label{human}
\end{figure*}

\begin{table}[h]
\centering
\caption{Error Analysis for a sample. Red text indicate observed performance deficiencies.}\label{error}
\scriptsize
\begin{tabular}{p{\columnwidth}}
\multicolumn{1}{c}{\includegraphics[scale = 0.3]{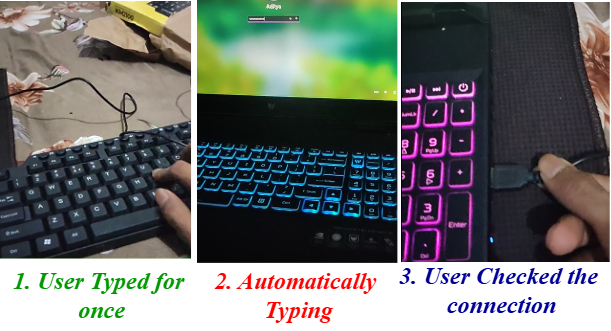}}\\
\hline
\textbf{Output from MR + VideoLLaMA2 -7b}: The user wants to complain that the keyboard is not functioning properly. Despite multiple attempts to press the keys, they remain unresponsive and instead automatically type other. {\color{red}{The user is disappointed with Amazon for selling defective items and is unable to contact customer care for a refund.}}\\
\hline
\textbf{Ground Truth}: The user wants to convey the complaint that  the keyboard is not working. The USB jack is also not working. Wrong typing occurs on the screen. words and trigger the escape button.\\
\hline
\end{tabular}
\end{table}

\subsection{Results and Discussions}
\paragraph{Answer to RQ1 Main Results:}
Evidently, the fine-tuned VideoLLaMA2-7b+MR model surpasses conventional baselines and state-of-the-art methods on the \textsc{CoD-V} task. As shown in Table~\ref{tab:ablation}, it consistently outperforms across both with emotion and without emotion settings, excelling in ROUGE, BLEU, BERTScore, and FRES, ensuring high fluency, contextual relevance, and nuanced emotional expression in complaint narratives. Compared to BLIP-VQA-BASE, QWEN2-VL-7b, Gemma3-12b, and LSTM-based models, it retains complaint-specific features more effectively, while lower PS, CLIS, and MoS scores indicate superior coherence and reduced semantic drift.

These gains stem from the Multimodal Retrieval-Augmented Generation (RAG) module, which dynamically incorporates relevant visual-textual evidence, yielding a 3–4\% improvement over fine-tuned VLMs alone. Unlike static architectures, RAG retrieves semantically aligned instances to improve factual grounding and visual-linguistic alignment, enabling fine-grained disambiguation (e.g., distinguishing typing errors from hardware faults). The MR further injects contextual cues (e.g., non-functionality, quality degradation) from the Amazon review dataset, enhancing the model’s ability to generalise across diverse complaint scenarios and reinforcing its robustness in \textsc{CoD-V}.

\paragraph{Answer to RQ2 Performance Observations:} Table \ref{tab:ablation} underscores the critical role of emotion consideration across all models. Emotion injection enhances contextual relevance and ensures a consistent tone, significantly improving the quality of complaint descriptions generated by the models. Notably, models incorporating emotion-based features outperformed the emotion-agnostic setup, highlighting the strength of context-specific model selection. Furthermore, we evaluated the models' performance through qualitative assessments, error analysis, and human evaluations, ensuring a thorough and unbiased validation process.\\
\textbf{\textit{1. Qualitative Analysis:}} Table \ref{qualitative} highlights the strengths of MR + VideoLLaMA2-7b, particularly its ability to generate complaints reflecting negative emotions. In the \textit{Without Emotion} category, MR + VideoLLaMA2-7b effectively captures the user's comprehensive concerns by detailing both the poor quality and the non-functionality of the mouse sensor, aligning closely with the ground truth, which emphasizes the product's defectiveness. In contrast, BLIP-VQA-BASE provides a basic complaint identification without addressing quality, while VideoLLaVA2-7b mentions the product's age but lacks specificity regarding functional issue aspects. In the \textit{With Emotion} category, VideoLLaMA2-7b+MR excels by articulating the \textit{user's frustration over the defective mouse and its poor quality}, showcasing a nuanced understanding of emotional expression. BLIP-VQA-BASE falls short by offering a superficial quality assessment, and VideoLLaVA2-7b acknowledges dissatisfaction regarding durability but lacks a direct emotional connection. Overall, VideoLLaMA2-7b+MR demonstrates superior capability in integrating detailed observations of product quality and user sentiment, positioning it as a more robust model compared to its counterparts.\\
\textbf{\textit{2. Error Example:}} Table \ref{error} shows an example where  
the MR + VideoLLaMA2-7b response introduces additional details not present in the ground truth, leading to an inaccurate representation of the user’s complaint. While the ground truth focuses on specific issues, namely, a non-functional keyboard, malfunctioning USB jack, and unintended typing on the screen, VideoLLaMA2-7b+MR includes information that the user is ``\textit{disappointed with Amazon}'' and ``\textit{unable to contact customer care for a refund}.'' These additions misinterpret the original complaint and could convey a sense of frustration or escalation not initially expressed by the user. These inaccuracies result in a response that overstates the situation, detracting from the precision required to represent the user’s original message.

\noindent \textbf{\textit{3. Human Evaluation:}}
For a comprehensive evaluation, we conducted a human assessment with {\em Category B} annotators on 80 randomly sampled test instances, using Information Preservation Ratings (IPR) across Adequacy, Fluency, Emotion Encoding, Relevance, and Visual Content Alignment. Scores ranged from 1 to 5, based on the extent of preservation. As shown in Figure.~\ref{human}, BLIP-VQA-BASE underperformed in Visual Content Alignment, while VideoLLaVA2-7b and VideoLLaMA2-7b+MR excelled. The proposed VideoLLaMA2-7b+MR outperformed both in emotion encoding and capturing nuanced emotional contexts more effectively. Average scores were cross-verified by {\em Category A} annotators for robustness.


 \paragraph{Answer to RQ3 (Uniqueness of {\em CoD-V} task):} The \textit{CoD-V} task differs significantly from traditional summary generation and video description tasks, as highlighted in Table \ref{task_difference}. We prompt multiple models for 3 tasks: Proposed \textsc{CoD-V}, summary generation (SG) and Video Description (VD). Unlike conventional tasks, which primarily rely on traditional lexical analysis metrics, the \textit{CoD-V} task introduces a unique focus on complaint-specific characteristics, such as the Vader Score, which is crucial for capturing the sentimental essence of complaints. This metric, along with the retention characteristics defined for complaint nature (CR), is central to the task. In \textit{CoD-V} task, the proposed VideoLLaMA2-7b+MR model has demonstrated better performance with more negative Vader Score, while traditional tasks such as summary generation and video description tend to yield more positive sentiment scores. Models such as GeminiFlash, GPT4O, and VideoLLaMA2-7b+MR show competitive results, but there remains a significant performance gap when evaluated on task-specific criteria. With \textit{CoD-V}  models consistently achieving high Complaint Retention (CR) scores, the complaint generation task clearly demonstrates its optimal alignment with the complaint mining domain while justifying Figure \ref{intro}.
\paragraph{Answer to RQ4 (Societal benefits and applicability):} The proposed work enhances video-to-text generation for complaint articulation, aiding less literate users in expressing grievances on e-commerce platforms such a Amazon. It processes unstructured video reviews to extract key issues (e.g., product defects), facilitating efficient resolutions. The VideoLLaMA2-7b+MR model, applied to \textit{CoD-V}, extends to diverse NLP tasks, including real-time patient issue reporting in healthcare, automated movie trailer summarization, and toxicity detection in online content, demonstrating broad applicability in structured text generation from multimodal inputs.

\section{Conclusion}
This paper presents \textit{Complaint Description from Videos (\textsc{CoD-V})}, aimed at addressing communication challenges faced by less literate consumers on e-commerce platforms. We introduce the \textsc{ComVID} dataset with 1,175 annotated complaint videos across 4 domains and 8 complaint aspects. A fine-tuned multimodal RAG-configured VideoLLaMA2-7b model is proposed for generating descriptive complaints, outperforming existing methods in transparency. To further validate the uniqueness of the \textsc{CoD-V} task, we introduce the Complaint Retention (CR) score, coupled with the Vader score (VS), distinguishing it from summarization and video description tasks. Additionally, emotion and non-emotion attribute prompting strategies are employed to enrich the diversity of complaint descriptions, ultimately improving customer service responsiveness for e-commerce.

\section{Acknowledgement}
The authors sincerely acknowledge the invaluable contributions of the dataset annotators, Tannu, Pavnee, and Jheel, for their consistent dedication and meticulous effort in the annotation process.
\section{GenAI Usage Disclosure}
Generative AI tools (ChatGPT, Perplexity) were used in a limited capacity, solely for minor grammatical refinement and formatting. The core research, including conceptualization, experimental design, implementation, analysis, and validation, was conducted entirely by the authors. All technical contributions, data curation, experiments, and interpretations remain the exclusive work of the authors.
\bibliographystyle{ACM-Reference-Format}
\bibliography{sample-base}


\begin{thebibliography}{41}


\ifx \showCODEN    \undefined \def \showCODEN     #1{\unskip}     \fi
\ifx \showISBNx    \undefined \def \showISBNx     #1{\unskip}     \fi
\ifx \showISBNxiii \undefined \def \showISBNxiii  #1{\unskip}     \fi
\ifx \showISSN     \undefined \def \showISSN      #1{\unskip}     \fi
\ifx \showLCCN     \undefined \def \showLCCN      #1{\unskip}     \fi
\ifx \shownote     \undefined \def \shownote      #1{#1}          \fi
\ifx \showarticletitle \undefined \def \showarticletitle #1{#1}   \fi
\ifx \showURL      \undefined \def \showURL       {\relax}        \fi
\providecommand\bibfield[2]{#2}
\providecommand\bibinfo[2]{#2}
\providecommand\natexlab[1]{#1}
\providecommand\showeprint[2][]{arXiv:#2}

\bibitem[Akin(2024)]%
        {akin2024enhancing}
\bibfield{author}{\bibinfo{person}{Mustafa~Seref Akin}.} \bibinfo{year}{2024}\natexlab{}.
\newblock \showarticletitle{Enhancing e-commerce competitiveness: A comprehensive analysis of customer experiences and strategies in the Turkish market}.
\newblock \bibinfo{journal}{\emph{Journal of Open Innovation: Technology, Market, and Complexity}} \bibinfo{volume}{10}, \bibinfo{number}{1} (\bibinfo{year}{2024}), \bibinfo{pages}{100222}.
\newblock


\bibitem[AlQahtani(2021)]%
        {alqahtani2021product}
\bibfield{author}{\bibinfo{person}{Arwa~SM AlQahtani}.} \bibinfo{year}{2021}\natexlab{}.
\newblock \showarticletitle{Product sentiment analysis for amazon reviews}.
\newblock \bibinfo{journal}{\emph{International Journal of Computer Science \& Information Technology (IJCSIT) Vol}}  \bibinfo{volume}{13} (\bibinfo{year}{2021}).
\newblock


\bibitem[Banerjee and Lavie(2005)]%
        {banerjee2005meteor}
\bibfield{author}{\bibinfo{person}{Satanjeev Banerjee} {and} \bibinfo{person}{Alon Lavie}.} \bibinfo{year}{2005}\natexlab{}.
\newblock \showarticletitle{METEOR: An automatic metric for MT evaluation with improved correlation with human judgments}. In \bibinfo{booktitle}{\emph{Proceedings of the acl workshop on intrinsic and extrinsic evaluation measures for machine translation and/or summarization}}. \bibinfo{pages}{65--72}.
\newblock


\bibitem[Coleman and Liau(1975)]%
        {coleman1975computer}
\bibfield{author}{\bibinfo{person}{Meri Coleman} {and} \bibinfo{person}{Ta~Lin Liau}.} \bibinfo{year}{1975}\natexlab{}.
\newblock \showarticletitle{A computer readability formula designed for machine scoring.}
\newblock \bibinfo{journal}{\emph{Journal of Applied Psychology}} \bibinfo{volume}{60}, \bibinfo{number}{2} (\bibinfo{year}{1975}), \bibinfo{pages}{283}.
\newblock


\bibitem[Das et~al\mbox{.}(2025)]%
        {das2025deciphering}
\bibfield{author}{\bibinfo{person}{Sarmistha Das}, \bibinfo{person}{Basha Mujavarsheik}, \bibinfo{person}{RE~Zera Lyngkhoi}, \bibinfo{person}{Sriparna Saha}, {and} \bibinfo{person}{Alka Maurya}.} \bibinfo{year}{2025}\natexlab{}.
\newblock \showarticletitle{Deciphering the complaint aspects: Towards an aspect-based complaint identification model with video complaint dataset in finance}. In \bibinfo{booktitle}{\emph{2025 IEEE/CVF Winter Conference on Applications of Computer Vision (WACV)}}. IEEE, \bibinfo{pages}{7195--7204}.
\newblock


\bibitem[Das et~al\mbox{.}(2023)]%
        {das2023let}
\bibfield{author}{\bibinfo{person}{Sarmistha Das}, \bibinfo{person}{Apoorva Singh}, \bibinfo{person}{Raghav Jain}, \bibinfo{person}{Sriparna Saha}, {and} \bibinfo{person}{Alka Maurya}.} \bibinfo{year}{2023}\natexlab{}.
\newblock \showarticletitle{Let the Model Make Financial Senses: A Text2Text Generative Approach for Financial Complaint Identification}. In \bibinfo{booktitle}{\emph{Pacific-Asia Conference on Knowledge Discovery and Data Mining}}. Springer, \bibinfo{pages}{58--69}.
\newblock


\bibitem[Das et~al\mbox{.}(2024a)]%
        {das2024negative}
\bibfield{author}{\bibinfo{person}{Sarmistha Das}, \bibinfo{person}{Apoorva Singh}, \bibinfo{person}{Sriparna Saha}, {and} \bibinfo{person}{Alka Maurya}.} \bibinfo{year}{2024}\natexlab{a}.
\newblock \showarticletitle{Negative review or complaint? Exploring interpretability in financial complaints}.
\newblock \bibinfo{journal}{\emph{Ieee Transactions on Computational Social Systems}} \bibinfo{volume}{11}, \bibinfo{number}{3} (\bibinfo{year}{2024}), \bibinfo{pages}{3606--3615}.
\newblock


\bibitem[Das et~al\mbox{.}(2024b)]%
        {10379488}
\bibfield{author}{\bibinfo{person}{Sarmistha Das}, \bibinfo{person}{Apoorva Singh}, \bibinfo{person}{Sriparna Saha}, {and} \bibinfo{person}{Alka Maurya}.} \bibinfo{year}{2024}\natexlab{b}.
\newblock \showarticletitle{Negative Review or Complaint? Exploring Interpretability in Financial Complaints}.
\newblock \bibinfo{journal}{\emph{IEEE Transactions on Computational Social Systems}} (\bibinfo{year}{2024}), \bibinfo{pages}{1--10}.
\newblock
\href{https://doi.org/10.1109/TCSS.2023.3338357}{doi:\nolinkurl{10.1109/TCSS.2023.3338357}}


\bibitem[Devanathan et~al\mbox{.}(2024)]%
        {devanathan2024seeing}
\bibfield{author}{\bibinfo{person}{Rishikesh Devanathan}, \bibinfo{person}{Apoorva Singh}, \bibinfo{person}{AS Poornash}, {and} \bibinfo{person}{Sriparna Saha}.} \bibinfo{year}{2024}\natexlab{}.
\newblock \showarticletitle{Seeing Beyond Words: Multimodal Aspect-Level Complaint Detection in Ecommerce Videos}. In \bibinfo{booktitle}{\emph{ACM Multimedia 2024}}.
\newblock


\bibitem[Farr et~al\mbox{.}(1951)]%
        {farr1951simplification}
\bibfield{author}{\bibinfo{person}{James~N Farr}, \bibinfo{person}{James~J Jenkins}, {and} \bibinfo{person}{Donald~G Paterson}.} \bibinfo{year}{1951}\natexlab{}.
\newblock \showarticletitle{Simplification of Flesch reading ease formula.}
\newblock \bibinfo{journal}{\emph{Journal of applied psychology}} \bibinfo{volume}{35}, \bibinfo{number}{5} (\bibinfo{year}{1951}), \bibinfo{pages}{333}.
\newblock


\bibitem[for Telecommunication~Sciences(2024)]%
        {its_video_quality_data}
\bibfield{author}{\bibinfo{person}{Institute for Telecommunication~Sciences}.} \bibinfo{year}{2024}\natexlab{}.
\newblock \bibinfo{title}{Video Quality Research Data}.
\newblock
\urldef\tempurl%
\url{https://its.ntia.gov/research/qoe/video-quality-research/data}
\showURL{%
\tempurl}
\newblock
\shownote{Accessed: 2024-11-01}.


\bibitem[He et~al\mbox{.}(2016)]%
        {he2016deep}
\bibfield{author}{\bibinfo{person}{Kaiming He}, \bibinfo{person}{Xiangyu Zhang}, \bibinfo{person}{Shaoqing Ren}, {and} \bibinfo{person}{Jian Sun}.} \bibinfo{year}{2016}\natexlab{}.
\newblock \showarticletitle{Deep residual learning for image recognition}. In \bibinfo{booktitle}{\emph{Proceedings of the IEEE conference on computer vision and pattern recognition}}. \bibinfo{pages}{770--778}.
\newblock


\bibitem[HELMY et~al\mbox{.}(2024)]%
        {helmy2024decision}
\bibfield{author}{\bibinfo{person}{YEHIA HELMY}, \bibinfo{person}{MERNA ASHRAF}, {and} \bibinfo{person}{LAILA ABDELHAMID}.} \bibinfo{year}{2024}\natexlab{}.
\newblock \showarticletitle{A DECISION SUPPORT MODEL TO IMPROVE COMPLAINT HANDLING IN E-COMMERCE TO ENHANCE CUSTOMER TRUST}.
\newblock \bibinfo{journal}{\emph{Journal of Theoretical and Applied Information Technology}} \bibinfo{volume}{102}, \bibinfo{number}{11} (\bibinfo{year}{2024}).
\newblock


\bibitem[Huang et~al\mbox{.}(2024)]%
        {huang2024research}
\bibfield{author}{\bibinfo{person}{Yanrong Huang}, \bibinfo{person}{Zhiyi He}, \bibinfo{person}{Han Lv}, {and} \bibinfo{person}{Jian Min}.} \bibinfo{year}{2024}\natexlab{}.
\newblock \showarticletitle{Research on Mining Negative Online Reviews on E-commerce Platforms Based on Social Network Analysis and LDA Model}. In \bibinfo{booktitle}{\emph{Intelligent Management of Data and Information in Decision Making: Proceedings of the 16th FLINS Conference on Computational Intelligence in Decision and Control \& the 19th ISKE Conference on Intelligence Systems and Knowledge Engineering (FLINS-ISKE 2024)}}. World Scientific, \bibinfo{pages}{177--185}.
\newblock


\bibitem[Hutto and Gilbert(2014)]%
        {hutto2014vader}
\bibfield{author}{\bibinfo{person}{Clayton Hutto} {and} \bibinfo{person}{Eric Gilbert}.} \bibinfo{year}{2014}\natexlab{}.
\newblock \showarticletitle{Vader: A parsimonious rule-based model for sentiment analysis of social media text}. In \bibinfo{booktitle}{\emph{Proceedings of the international AAAI conference on web and social media}}, Vol.~\bibinfo{volume}{8}. \bibinfo{pages}{216--225}.
\newblock


\bibitem[Jain et~al\mbox{.}(2023)]%
        {jain2023abcord}
\bibfield{author}{\bibinfo{person}{Raghav Jain}, \bibinfo{person}{Apoorva Singh}, \bibinfo{person}{Vivek Gangwar}, {and} \bibinfo{person}{Sriparna Saha}.} \bibinfo{year}{2023}\natexlab{}.
\newblock \showarticletitle{AbCoRD: Exploiting multimodal generative approach for Aspect-based Complaint and Rationale Detection}. In \bibinfo{booktitle}{\emph{Proceedings of the 31st ACM International Conference on Multimedia}}. \bibinfo{pages}{8571--8579}.
\newblock


\bibitem[Jelinek et~al\mbox{.}(1977)]%
        {jelinek1977perplexity}
\bibfield{author}{\bibinfo{person}{Fred Jelinek}, \bibinfo{person}{Robert~L Mercer}, \bibinfo{person}{Lalit~R Bahl}, {and} \bibinfo{person}{James~K Baker}.} \bibinfo{year}{1977}\natexlab{}.
\newblock \showarticletitle{Perplexity—a measure of the difficulty of speech recognition tasks}.
\newblock \bibinfo{journal}{\emph{The Journal of the Acoustical Society of America}} \bibinfo{volume}{62}, \bibinfo{number}{S1} (\bibinfo{year}{1977}), \bibinfo{pages}{S63--S63}.
\newblock


\bibitem[Jin and Aletras(2021)]%
        {jin2021modeling}
\bibfield{author}{\bibinfo{person}{Mali Jin} {and} \bibinfo{person}{Nikolaos Aletras}.} \bibinfo{year}{2021}\natexlab{}.
\newblock \showarticletitle{Modeling the severity of complaints in social media}.
\newblock \bibinfo{journal}{\emph{arXiv preprint arXiv:2103.12428}} (\bibinfo{year}{2021}).
\newblock


\bibitem[Johnson et~al\mbox{.}(2019)]%
        {johnson2019billion}
\bibfield{author}{\bibinfo{person}{Jeff Johnson}, \bibinfo{person}{Matthijs Douze}, {and} \bibinfo{person}{Herv{\'e} J{\'e}gou}.} \bibinfo{year}{2019}\natexlab{}.
\newblock \showarticletitle{Billion-scale similarity search with GPUs}.
\newblock \bibinfo{journal}{\emph{IEEE Transactions on Big Data}} \bibinfo{volume}{7}, \bibinfo{number}{3} (\bibinfo{year}{2019}), \bibinfo{pages}{535--547}.
\newblock


\bibitem[Li et~al\mbox{.}(2023)]%
        {li2023blip2}
\bibfield{author}{\bibinfo{person}{Junnan Li}, \bibinfo{person}{Dongxu Li}, \bibinfo{person}{Silvio Savarese}, {and} \bibinfo{person}{Steven Hoi}.} \bibinfo{year}{2023}\natexlab{}.
\newblock \showarticletitle{{BLIP-2:} Bootstrapping Language-Image Pre-training with Frozen Image Encoders and Large Language Models}. In \bibinfo{booktitle}{\emph{ICML}}.
\newblock


\bibitem[Li et~al\mbox{.}(2017)]%
        {li2017reader}
\bibfield{author}{\bibinfo{person}{Piji Li}, \bibinfo{person}{Lidong Bing}, {and} \bibinfo{person}{Wai Lam}.} \bibinfo{year}{2017}\natexlab{}.
\newblock \showarticletitle{Reader-Aware Multi-Document Summarization: An Enhanced Model and The First Dataset}. In \bibinfo{booktitle}{\emph{Proceedings of the Workshop on New Frontiers in Summarization}}. \bibinfo{pages}{91--99}.
\newblock


\bibitem[Lin(2004)]%
        {lin-2004-rouge}
\bibfield{author}{\bibinfo{person}{Chin-Yew Lin}.} \bibinfo{year}{2004}\natexlab{}.
\newblock \showarticletitle{{ROUGE}: A Package for Automatic Evaluation of Summaries}. In \bibinfo{booktitle}{\emph{Text Summarization Branches Out}}. \bibinfo{publisher}{Association for Computational Linguistics}, \bibinfo{address}{Barcelona, Spain}, \bibinfo{pages}{74--81}.
\newblock
\urldef\tempurl%
\url{https://aclanthology.org/W04-1013}
\showURL{%
\tempurl}


\bibitem[Lin and Och(2004)]%
        {lin-och-2004-orange}
\bibfield{author}{\bibinfo{person}{Chin-Yew Lin} {and} \bibinfo{person}{Franz~Josef Och}.} \bibinfo{year}{2004}\natexlab{}.
\newblock \showarticletitle{{ORANGE}: a Method for Evaluating Automatic Evaluation Metrics for Machine Translation}. In \bibinfo{booktitle}{\emph{{COLING} 2004: Proceedings of the 20th International Conference on Computational Linguistics}}. \bibinfo{publisher}{COLING}, \bibinfo{address}{Geneva, Switzerland}, \bibinfo{pages}{501--507}.
\newblock
\urldef\tempurl%
\url{https://www.aclweb.org/anthology/C04-1072}
\showURL{%
\tempurl}


\bibitem[Liu et~al\mbox{.}(2021)]%
        {liu2021cma}
\bibfield{author}{\bibinfo{person}{Huidong Liu}, \bibinfo{person}{Shaoyuan Xu}, \bibinfo{person}{Jinmiao Fu}, \bibinfo{person}{Yang Liu}, \bibinfo{person}{Ning Xie}, \bibinfo{person}{Chien-Chih Wang}, \bibinfo{person}{Bryan Wang}, {and} \bibinfo{person}{Yi Sun}.} \bibinfo{year}{2021}\natexlab{}.
\newblock \showarticletitle{Cma-clip: Cross-modality attention clip for image-text classification}.
\newblock \bibinfo{journal}{\emph{arXiv preprint arXiv:2112.03562}} (\bibinfo{year}{2021}).
\newblock


\bibitem[Poria et~al\mbox{.}(2018)]%
        {poria2018meld}
\bibfield{author}{\bibinfo{person}{Soujanya Poria}, \bibinfo{person}{Devamanyu Hazarika}, \bibinfo{person}{Navonil Majumder}, \bibinfo{person}{Gautam Naik}, \bibinfo{person}{Erik Cambria}, {and} \bibinfo{person}{Rada Mihalcea}.} \bibinfo{year}{2018}\natexlab{}.
\newblock \showarticletitle{Meld: A multimodal multi-party dataset for emotion recognition in conversations}.
\newblock \bibinfo{journal}{\emph{arXiv preprint arXiv:1810.02508}} (\bibinfo{year}{2018}).
\newblock


\bibitem[Preotiuc{-}Pietro et~al\mbox{.}(2019)]%
        {DBLP:conf/acl/Preotiuc-Pietro19a}
\bibfield{author}{\bibinfo{person}{Daniel Preotiuc{-}Pietro}, \bibinfo{person}{Mihaela Gaman}, {and} \bibinfo{person}{Nikolaos Aletras}.} \bibinfo{year}{2019}\natexlab{}.
\newblock \showarticletitle{Automatically Identifying Complaints in Social Media}. In \bibinfo{booktitle}{\emph{Proceedings of the 57th Conference of the Association for Computational Linguistics, {ACL} 2019, Florence, Italy, July 28- August 2, 2019, Volume 1: Long Papers}}, \bibfield{editor}{\bibinfo{person}{Anna Korhonen}, \bibinfo{person}{David~R. Traum}, {and} \bibinfo{person}{Llu{\'{\i}}s M{\`{a}}rquez}} (Eds.). \bibinfo{publisher}{Association for Computational Linguistics}, \bibinfo{pages}{5008--5019}.
\newblock
\href{https://doi.org/10.18653/v1/p19-1495}{doi:\nolinkurl{10.18653/v1/p19-1495}}


\bibitem[Radford et~al\mbox{.}(2021)]%
        {radford2021learning}
\bibfield{author}{\bibinfo{person}{Alec Radford}, \bibinfo{person}{Jong~Wook Kim}, \bibinfo{person}{Chris Hallacy}, \bibinfo{person}{Aditya Ramesh}, \bibinfo{person}{Gabriel Goh}, \bibinfo{person}{Sandhini Agarwal}, \bibinfo{person}{Girish Sastry}, \bibinfo{person}{Amanda Askell}, \bibinfo{person}{Pamela Mishkin}, \bibinfo{person}{Jack Clark}, {et~al\mbox{.}}} \bibinfo{year}{2021}\natexlab{}.
\newblock \showarticletitle{Learning transferable visual models from natural language supervision}. In \bibinfo{booktitle}{\emph{International conference on machine learning}}. PMLR, \bibinfo{pages}{8748--8763}.
\newblock


\bibitem[Roy et~al\mbox{.}(2021)]%
        {roy2021attribute}
\bibfield{author}{\bibinfo{person}{Kalyani Roy}, \bibinfo{person}{Pawan Goyal}, {and} \bibinfo{person}{Manish Pandey}.} \bibinfo{year}{2021}\natexlab{}.
\newblock \showarticletitle{Attribute value generation from product title using language models}. In \bibinfo{booktitle}{\emph{Proceedings of The 4th Workshop on e-Commerce and NLP}}. \bibinfo{pages}{13--17}.
\newblock


\bibitem[Saha et~al\mbox{.}(2021)]%
        {saha2021towards}
\bibfield{author}{\bibinfo{person}{Tulika Saha}, \bibinfo{person}{Apoorva Upadhyaya}, \bibinfo{person}{Sriparna Saha}, {and} \bibinfo{person}{Pushpak Bhattacharyya}.} \bibinfo{year}{2021}\natexlab{}.
\newblock \showarticletitle{Towards sentiment and emotion aided multi-modal speech act classification in twitter}. In \bibinfo{booktitle}{\emph{Proceedings of the 2021 conference of the North American chapter of the association for computational linguistics: Human language technologies}}. \bibinfo{pages}{5727--5737}.
\newblock


\bibitem[Sesver et~al\mbox{.}(2022)]%
        {sesver2022vidi}
\bibfield{author}{\bibinfo{person}{Duygu Sesver}, \bibinfo{person}{Alp~Eren Gen{\c{c}}o{\u{g}}lu}, \bibinfo{person}{{\c{C}}a{\u{g}}r{\i}~Emre Y{\i}ld{\i}z}, \bibinfo{person}{Zehra G{\"u}nindi}, \bibinfo{person}{Faeze Habibi}, \bibinfo{person}{Ziya~Ata Yaz{\i}c{\i}}, {and} \bibinfo{person}{Haz{\i}m~Kemal Ekenel}.} \bibinfo{year}{2022}\natexlab{}.
\newblock \showarticletitle{VIDI: A Video Dataset of Incidents}. In \bibinfo{booktitle}{\emph{2022 IEEE 14th Image, Video, and Multidimensional Signal Processing Workshop (IVMSP)}}. IEEE, \bibinfo{pages}{1--5}.
\newblock


\bibitem[Simonyan and Zisserman(2014)]%
        {simonyan2014very}
\bibfield{author}{\bibinfo{person}{Karen Simonyan} {and} \bibinfo{person}{Andrew Zisserman}.} \bibinfo{year}{2014}\natexlab{}.
\newblock \showarticletitle{Very deep convolutional networks for large-scale image recognition}.
\newblock \bibinfo{journal}{\emph{arXiv preprint arXiv:1409.1556}} (\bibinfo{year}{2014}).
\newblock


\bibitem[Singh et~al\mbox{.}(2022a)]%
        {singh2022sentiment}
\bibfield{author}{\bibinfo{person}{Apoorva Singh}, \bibinfo{person}{Soumyodeep Dey}, \bibinfo{person}{Anamitra Singha}, {and} \bibinfo{person}{Sriparna Saha}.} \bibinfo{year}{2022}\natexlab{a}.
\newblock \showarticletitle{Sentiment and emotion-aware multi-modal complaint identification}. In \bibinfo{booktitle}{\emph{Proceedings of the AAAI Conference on Artificial Intelligence}}, Vol.~\bibinfo{volume}{36}. \bibinfo{pages}{12163--12171}.
\newblock


\bibitem[Singh et~al\mbox{.}(2023)]%
        {singh2023knowing}
\bibfield{author}{\bibinfo{person}{Apoorva Singh}, \bibinfo{person}{Vivek Gangwar}, \bibinfo{person}{Shubham Sharma}, {and} \bibinfo{person}{Sriparna Saha}.} \bibinfo{year}{2023}\natexlab{}.
\newblock \showarticletitle{Knowing what and how: a multi-modal aspect-based framework for complaint detection}. In \bibinfo{booktitle}{\emph{European Conference on Information Retrieval}}. Springer, \bibinfo{pages}{125--140}.
\newblock


\bibitem[Singh et~al\mbox{.}(2022b)]%
        {singh2022adversarial}
\bibfield{author}{\bibinfo{person}{Apoorva Singh}, \bibinfo{person}{Arousha Nazir}, {and} \bibinfo{person}{Sriparna Saha}.} \bibinfo{year}{2022}\natexlab{b}.
\newblock \showarticletitle{Adversarial Multi-task Model for Emotion, Sentiment, and Sarcasm Aided Complaint Detection}. In \bibinfo{booktitle}{\emph{Advances in Information Retrieval: 44th European Conference on IR Research, ECIR 2022, Stavanger, Norway, April 10--14, 2022, Proceedings, Part I}}. Springer, \bibinfo{pages}{428--442}.
\newblock


\bibitem[Team(2025)]%
        {gemma_2025}
\bibfield{author}{\bibinfo{person}{Gemma Team}.} \bibinfo{year}{2025}\natexlab{}.
\newblock \showarticletitle{Gemma 3}.
\newblock  (\bibinfo{year}{2025}).
\newblock
\urldef\tempurl%
\url{https://goo.gle/Gemma3Report}
\showURL{%
\tempurl}


\bibitem[Wang et~al\mbox{.}(2024)]%
        {Qwen2VL}
\bibfield{author}{\bibinfo{person}{Peng Wang}, \bibinfo{person}{Shuai Bai}, \bibinfo{person}{Sinan Tan}, \bibinfo{person}{Shijie Wang}, \bibinfo{person}{Zhihao Fan}, \bibinfo{person}{Jinze Bai}, \bibinfo{person}{Keqin Chen}, \bibinfo{person}{Xuejing Liu}, \bibinfo{person}{Jialin Wang}, \bibinfo{person}{Wenbin Ge}, \bibinfo{person}{Yang Fan}, \bibinfo{person}{Kai Dang}, \bibinfo{person}{Mengfei Du}, \bibinfo{person}{Xuancheng Ren}, \bibinfo{person}{Rui Men}, \bibinfo{person}{Dayiheng Liu}, \bibinfo{person}{Chang Zhou}, \bibinfo{person}{Jingren Zhou}, {and} \bibinfo{person}{Junyang Lin}.} \bibinfo{year}{2024}\natexlab{}.
\newblock \showarticletitle{Qwen2-VL: Enhancing Vision-Language Model's Perception of the World at Any Resolution}.
\newblock \bibinfo{journal}{\emph{arXiv preprint arXiv:2409.12191}} (\bibinfo{year}{2024}).
\newblock


\bibitem[Wassan et~al\mbox{.}(2021)]%
        {wassan2021amazon}
\bibfield{author}{\bibinfo{person}{Sobia Wassan}, \bibinfo{person}{Xi Chen}, \bibinfo{person}{Tian Shen}, \bibinfo{person}{Muhammad Waqar}, {and} \bibinfo{person}{NZ Jhanjhi}.} \bibinfo{year}{2021}\natexlab{}.
\newblock \showarticletitle{Amazon product sentiment analysis using machine learning techniques}.
\newblock \bibinfo{journal}{\emph{Revista Argentina de Cl{\'\i}nica Psicol{\'o}gica}} \bibinfo{volume}{30}, \bibinfo{number}{1} (\bibinfo{year}{2021}), \bibinfo{pages}{695}.
\newblock


\bibitem[Xu et~al\mbox{.}(2022)]%
        {xu2022gmflow}
\bibfield{author}{\bibinfo{person}{Haofei Xu}, \bibinfo{person}{Jing Zhang}, \bibinfo{person}{Jianfei Cai}, \bibinfo{person}{Hamid Rezatofighi}, {and} \bibinfo{person}{Dacheng Tao}.} \bibinfo{year}{2022}\natexlab{}.
\newblock \showarticletitle{Gmflow: Learning optical flow via global matching}. In \bibinfo{booktitle}{\emph{Proceedings of the IEEE/CVF conference on computer vision and pattern recognition}}. \bibinfo{pages}{8121--8130}.
\newblock


\bibitem[Zhang et~al\mbox{.}(2019)]%
        {zhang2019bertscore}
\bibfield{author}{\bibinfo{person}{Tianyi Zhang}, \bibinfo{person}{Varsha Kishore}, \bibinfo{person}{Felix Wu}, \bibinfo{person}{Kilian~Q Weinberger}, {and} \bibinfo{person}{Yoav Artzi}.} \bibinfo{year}{2019}\natexlab{}.
\newblock \showarticletitle{Bertscore: Evaluating text generation with bert}.
\newblock \bibinfo{journal}{\emph{arXiv preprint arXiv:1904.09675}} (\bibinfo{year}{2019}).
\newblock


\bibitem[Zhao et~al\mbox{.}(2019)]%
        {zhao2019moverscore}
\bibfield{author}{\bibinfo{person}{Wei Zhao}, \bibinfo{person}{Maxime Peyrard}, \bibinfo{person}{Fei Liu}, \bibinfo{person}{Yang Gao}, \bibinfo{person}{Christian~M. Meyer}, {and} \bibinfo{person}{Steffen Eger}.} \bibinfo{year}{2019}\natexlab{}.
\newblock \showarticletitle{MoverScore: Text Generation Evaluating with Contextualized Embeddings and Earth Mover Distance}. In \bibinfo{booktitle}{\emph{Proceedings of the 2019 Conference on Empirical Methods in Natural Language Processing}}. \bibinfo{publisher}{Association for Computational Linguistics}, \bibinfo{address}{Hong Kong, China}.
\newblock


\bibitem[Zhu et~al\mbox{.}(2023)]%
        {zhu2023languagebind}
\bibfield{author}{\bibinfo{person}{Bin Zhu}, \bibinfo{person}{Bin Lin}, \bibinfo{person}{Munan Ning}, \bibinfo{person}{Yang Yan}, \bibinfo{person}{Jiaxi Cui}, \bibinfo{person}{HongFa Wang}, \bibinfo{person}{Yatian Pang}, \bibinfo{person}{Wenhao Jiang}, \bibinfo{person}{Junwu Zhang}, \bibinfo{person}{Zongwei Li}, {et~al\mbox{.}}} \bibinfo{year}{2023}\natexlab{}.
\newblock \showarticletitle{LanguageBind: Extending Video-Language Pretraining to N-modality by Language-based Semantic Alignment}.
\newblock \bibinfo{journal}{\emph{arXiv preprint arXiv:2310.01852}} (\bibinfo{year}{2023}).
\newblock


\end{thebibliography}

\appendix









\end{document}